\documentclass{article} 
\usepackage{iclr2025_conference,times}
\usepackage[pdftex]{graphicx}
\usepackage{multirow}
\usepackage{booktabs}

\usepackage{xcolor}
\usepackage{soul}
\usepackage{hyperref}
\usepackage{caption}
\usepackage{url}
\usepackage{listings}
\usepackage{wrapfig}
\usepackage{amsmath,amsfonts,amssymb}
\usepackage{algorithm,algorithmic}

\usepackage{amsmath,amsfonts,bm}

\def\Figref#1{Figure~\ref{#1}}
\def\Tabref#1{Table~\ref{#1}}

\def\eqref#1{equation~\ref{#1}}

\def\1{\bm{1}}

\def\vt{{\bm{t}}}

\def\vw{{\bm{w}}}

\def\mI{{\bm{I}}}

\def\mM{{\bm{M}}}

\def\mU{{\bm{U}}}

\DeclareMathAlphabet{\mathsfit}{\encodingdefault}{\sfdefault}{m}{sl}
\SetMathAlphabet{\mathsfit}{bold}{\encodingdefault}{\sfdefault}{bx}{n}

\def\sD{{\mathbb{D}}}

\def\sT{{\mathbb{T}}}

\DeclareMathOperator*{\argmax}{arg\,max}
\DeclareMathOperator*{\argmin}{arg\,min}

\usepackage{hyperref}
\usepackage{url}
\iclrfinalcopy

\title{Optimizing Pretraining Data Mixtures \\ with LLM-Estimated Utility}

\author{William Held\thanks {Work completed during an internship at Meta AI. Contact: held@stanford.edu, tbmihaylov@meta.com}$\texttt{ }^{\sigma,\gamma}$ \quad Bhargavi Paranjape$^{\mu}$  \quad Punit Singh Koura$^{\mu}$ \\ \textbf{Mike Lewis$^{\mu}$  \quad Frank Zhang$^{\mu}$  \quad Todor Mihaylov$^{\mu}$}\\
$^\mu$Meta AI \quad $^{\sigma}$Stanford University \quad $^{\gamma}$Georgia Institute of Technology
}

\begin{document}

\maketitle

\begin{abstract}
Large Language Models improve with increasing amounts of high-quality training data. However, leveraging larger datasets requires balancing quality, quantity, and diversity across sources. After evaluating nine baseline methods under both compute- and data-constrained scenarios, we find token-count heuristics outperform manual and learned mixes, indicating that simple approaches accounting for dataset size and diversity are surprisingly effective. Building on this insight, we propose two complementary approaches: UtiliMax, which extends token-based heuristics by incorporating utility estimates from reduced-scale ablations, achieving up to a 10.6x speedup over manual baselines; and Model Estimated Data Utility (MEDU), which leverages LLMs to estimate data utility from small samples, matching ablation-based performance while reducing computational requirements by $\sim$200x\footnote{Comparison between a training run using $6\times 10^{19}$ FLOPs and inference cost of $3\times 10^{17}$ FLOPs from Llama 70B on 2.1 million tokens needed for MEDU using the FLOP equations from \citet{ComputeOptimal}}. Together, these approaches establish a new framework for automated, compute-efficient data mixing that is robust across training regimes.
\end{abstract}

\vspace{-1em}
\begin{figure}[h]
\begin{center}
\centering
\includegraphics[width=0.9\linewidth]{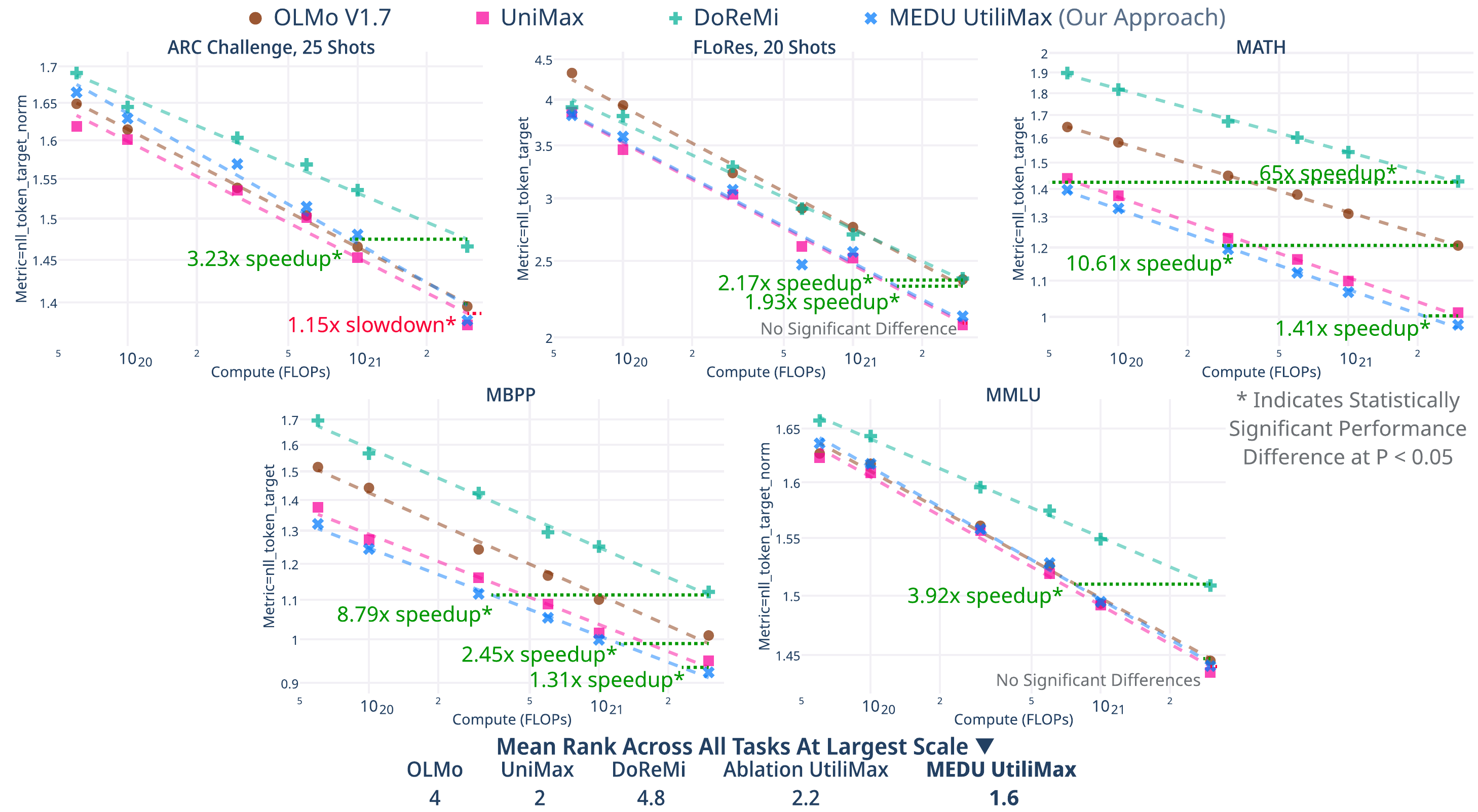}
\end{center}
\caption{Scaling curves for data mixing methods for compute-optimal models trained for $6 \times 10^{19}$ to $3 \times 10^{21}$ Floating Point Operations (FLOPs). Compared to manual~\citep[OLMo]{OLMo}, heuristic~\citep[UniMax]{UniMax}, and learned~\citep[DoReMi]{DoReMi} data mixes, UtiliMax leads to more compute efficient models that perform better on average across tasks.}
\end{figure}
\vspace{-1em}

\section{Introduction}

Large Language Model (LLM) pretraining data increasingly consists of sub-corpora from many sources covering multiple domains and varying in size~\citep{Pile, GLaM, RedPajama, ROOTS, Llama3}. Unlike traditional multi-task learning scenarios, datasets are not necessarily aligned with a specific intended use. Moreover, ``intended usage" is often multi-functional as LLMs are being developed for general-purpose functionality~\citep{GPTareGPT, LLMsAreGeneralPurpose}. Given multiple training corpora and multiple downstream goals, how should we sample from each corpus to get the best possible model?

Prior work has explored heuristic~\citep{Gopher, Dolma} and learned~\citep{DoReMi, ODM} approaches to solve this. However, there is minimal comparison between these methods using the same data and model configuration. Furthermore, it is unclear whether these approaches are robust to the impacts of epoching which is critical as frontier models are increasingly data-constrained~\citep{EpochAIData, StateOfDataConsent}. We address both questions by training nine baselines across six compute scales, evaluating them under constrained and unconstrained data scenarios. We then propose a novel method to optimize data mixes for a target budget based on the Markowitz model of portfolio optimization~\citep{OGPort, MarkAt70}.

\textbf{Research Questions \& Findings}
\begin{enumerate}
    \item \textit{How well do prior data mixing methods perform across compute scales and token budgets?} We re-implement and compare nine baselines in a unified setup across likely training scenarios. We find UniMax~\citep{UniMax}, an approach which maximizes diversity under epoching constraints, outperforms other heuristic, manual, and learned data mixes.
    
    \item \textit{Given a set of training runs on individual datasets, can we optimize a data mix effectively?} We propose UtiliMax, which combines utility estimates and dataset size to find data mixes using portfolio optimization. We show UtiliMax improves results by using reduced-scale ablations on individual datasets to estimate data utility. 
    
    \item \textit{Can we further reduce the cost of data mixing using existing LLMs to estimate utility?} We prompt LLMs to describe useful training data based on benchmark development sets, then use these descriptions to classify utility for a sample of the training data. These estimates are effective utility estimates for UtiliMax, but are 200x less costly to compute.
\end{enumerate}

\section{Background} \label{sec:background}
Given an arbitrary utility function $\mu$, data mix optimization aims to improve model performance by optimizing sampling weights $\vw$ over datasets $\sD := \{d_1, d_2, \ldots, d_n\}$. Based on external constraints such as computational resources or data, models are trained for a fixed budget $B_e$, in terms of a number of training examples. Given a budget, we want to find $\argmax_{\vw} \mu\left(\bigcup_{d \in \sD} \{e_1, e_2, \ldots, e_{w_d \cdot B_e} \} \sim d \right)$ where $e_i$ represents an example from dataset $d_i$.

Clearly, the target budget is a core factor for this problem, as well as data curation broadly~\citep{DFScalingLaws}. However, existing works run experiments at a single scale and budget~\citep{Gopher, Dolma, DoReMi, ODM}. To determine data mixes for the varying constraints of new models, data mixing methods must perform well across budgets and model scales. This makes evaluating this form of generalization critical for identifying effective methods.

With budget in mind, data mixing is a resource allocation problem. While resource allocation has been extensively studied in convex optimization~\citep{CVXBook}, optimizing utility $\mu$ directly is intractable given the cost of full-scale training runs. Furthermore, benchmarks only estimate $\mu$ based on a sample of the domain they represent. Instead, data mix optimization requires methods that are robust to error induced by \textit{estimated} utility.

Most relevant is the Markowitz model~\citep{OGPort}, a portfolio optimization method that balances an expected reward $\vw^\intercal\mu(\sD)$ with risk $\vw^\intercal\Sigma\vw$, where $\Sigma$ is the covariance matrix across assets. Using this \textit{risk-adjusted return}, data mix optimization can be formulated as:

\begin{equation}
    \argmax_{\vw} \vw^{\intercal}\mu(\sD) - \vw^\intercal\Sigma\vw \quad \text{subject to} \quad \pmb{1}^\intercal\vw = 1, \texttt{ min}(\vw) > 0,
    \label{eq:abstract}
\end{equation}

In applying this approach to data mix optimization, we make a few key assumptions. First, we assume the utility contributed by a dataset is linear with respect to its weight in the data mix. Second, we assume that the utility of each dataset is independent. Finally, to avoid compounding estimation error, we approximate $\Sigma$ as $|\sD|\mI$, corresponding to an assumption that the risk associated with each individual dataset increases linearly with the total number of alternatives.

\subsection{Related Work}\label{sec:related}

\paragraph{Token-Size Heuristic Data Mixes} Data mixes are often found using heuristics using the number of tokens per dataset $\vt$, the target budget $B_t$ in terms of tokens, and the sampling proportion $\frac{B_t\cdot\vw}{\vt}$. 

\textit{Uniform sampling} is the simplest baseline data mixing method and defines $\vw = \frac{1}{|\sD|}$. Despite the simplicity, uniform sampling is a strong baseline in multi-task learning~\citep{MultiTaskUniform}. 

\textit{Proportional sampling} is more common in efforts for large-scale training runs~\citep{Gopher, OLMo} and defines $\vw = \frac{\vt}{\vt^\intercal\pmb{1}}$. This holds the sampling proportion constant across datasets at any budget, minimizing the maximum sampling proportion or number of epochs. 

\textit{OLMo V1.7} utilizes near proportional weights, with Wikipedia up-sampled and CommonCrawl data down-sampled -- both by a factor of two\footnote{Information drawn from the \href{https://blog.allenai.org/olmo-1-7-7b-a-24-point-improvement-on-mmlu-92b43f7d269d}{OLMo V1.7 Release} blog post.}. A methodology for these adjustments is not released, but likely stems from a combination of researcher intuition and results from the data mix ablations shown in \citet{Dolma}. Since we use Dolma V1.7, we compare to this as a manual baseline.

\textit{UniMax}~\citep{UniMax} interpolates between proportional and uniform sampling by setting an epoch cap $C$ and finding $\argmin_{\vw} \vw^T\vw\text{ s.t. } \frac{B_T\cdot\vw}{\vt} \leq C$. Through the lens of portfolio optimization, UniMax purely minimizes risk under an assumption of uniform ``linguistic utility" from the multilingual setting it was designed for~\citep{LingUtil}. UtiliMax is a generalization of UniMax which allows for arbitrary non-uniform utility functions over datasets.

\paragraph{Learning Data Mixes}
There is also significant appeal to methods that \textit{learn} data mixes.

\textit{DoReMi}~\citep{DoReMi} does this using a sequence of training runs. First, a reference model is trained using a prior data mix. Then, a proxy model is trained using Distributionally Robust Optimization~\citep{DRO} to find weights which minimize worst-case excess loss with respect to the reference model. $\vw$ is defined as the average of these weights throughout training.

\textit{Online Data Mixing}~\citep[ODM]{ODM} treats data mixing as a multi-armed bandit problem and uses a variant of the EXP3 algorithm~\citep{EXP3} to dynamically sample data during training. Bandit methods are another natural formulation of data mixing, and have also been explored in works for multilinguality~\citep{FAIR} and translation~\citep{DontFollowRules}.

\paragraph{Model Based Quality Filtering} Related to MEDU, many prior works develop methods to filter out ``low-quality" data points entirely. \citet{SelectionSurvey} offer a systematic survey of this area.

\textit{Perplexity filters} use n-gram or other low-cost language models, such as KenLM~\citep{KenLM}, trained on high-quality text to assess data quality in new data~\citep{CCNet}. High perplexity data points are excluded based on the assumption that they are likely not natural language.

\textit{Quality classifiers} operate in a similar manner, but model both low and high quality data to distinguish the two. \citet{GPT3} popularized this approach by using a classifier trained to distinguish high-quality web pages from random web pages. Recently, LLMs have been used to create zero-shot quality classifiers based on natural language specifications of high-quality data~\citep{QuRating, FineWeb}. This approach has been validated at frontier model scale by Llama 3~\citep{Llama3}, but requires a single manually-written specification of ``high-quality" data.

\paragraph{How UtiliMax Differs} 

Pior data mixing work avoids making assumptions about use-cases to improve generality. On the other hand, most practitioners have a set of \textit{intended} use-cases measured by benchmarks which have strong correlation with various LLM capabilities~\citep{Observational}. UtiliMax maintains generality by optimizing for \textit{multiple} downstream tasks with terms for data utility, diversity, and size. This is applicable to any estimator, such as concurrent work which identifies loss correlations across open-source models~\citep{CorrelTristan}. Separately, MEDU proposes an approach to automatically construct quality specifications for each downstream task and then leverages this specification to provide more compute efficient utility estimates to UtiliMax.

\section{Experimental Setup}
\subsection{Training Setup}
\paragraph{Training Data Overview} We use Dolma V1.7~\citep{Dolma}, which is released under the Open Data Commons License, for our experiments. While prior works have used the Pile~\citep{Pile} for data mixing experiments, it has since had sections removed due to copyright issues which prevents direct comparison. Dolma is made up of 15 separate corpora including 2 corpora which are bucketed at higher granularity using KenLM~\citep{KenLM} perplexity. We report the names and sizes of the Dolma corpora using the Llama tokenizer in \Tabref{tab:dolma-stats}.

Importantly, Dolma is large-enough for and has been validated through large-scale training runs through OLMo~\citep{OLMo}. Dolma has also undergone document filtering, deduplication, and cleaning, which allows this work to focus solely on \textit{mixing} similarly preprocessed corpora.
 
\begin{table}[t]
\caption{Training Corpora Statistics From Dolma V1.7 using the Llama 3 Tokenizer.}
\label{tab:dolma-stats}
\begin{center}
\resizebox{0.85\linewidth}{!}{%
\begin{tabular}{cccccc}
\toprule
Corpus         & Tokens     & Corpus           & Tokens    & Corpus                 & Tokens                        \\ \cmidrule(l){1-2} \cmidrule(lr){3-4} \cmidrule(lr){5-6}
Refined Web     & 440B & PeS2o            & 58B & CC News Head         & 8.5B                     \\
CC Head   & 346B & Arxiv            & 27B & CC News Middle       & 3.7B                      \\
CC Middle & 436B & StackExchange    & 17B & CC News Tail          & 1.5B                      \\
CC Tail   & 371B & Tulu Flan       & 13B& MegaWika      & 4.4B                      \\
StarCoder      & 215B & Algebraic Stack & 11B & Wiki                   & 3.7B                     \\
\cline{5-6}
C4             & 133B & Open Web Math  & 5.1B  & \multirow{2}{*}{\textbf{Total}} & \multirow{2}{*}{\textbf{2.1T}} \\
Reddit         & 76B  & Books            & 5B  &                        &         \\                          \bottomrule
\end{tabular}
}
\end{center}
\end{table}

\paragraph{General Hyperparameters}
We train compute-optimal models from $6 \times 10^{19}$ to $3 \times 10^{21}$ FLOPs based on the scaling law presented in the Llama 3 paper~\citep{Llama3}, using the same architecture and tokenizer. The models range in size from 550M to 4.1B parameters and are trained on 14B to 110B tokens across compute scales. Across all training runs, we use a cosine learning rate schedule with a linear warmup for 2,000 training steps decaying to 0.1 of the peak learning rate. The peak learning rate is $2 \times 10^{-4}$ for all models, except for the largest run which uses $3\times10^{-4}$. Examples are packed to a sequence length of 8192 and batch size increases from 32 to 256 such that models train for approximately the same number of steps (58k on average, $\pm$9.2k)\footnote{We describe how examples are constructed, shuffled, and sampled in \ref{app:data_loader}.}.

\subsection{Evaluating Across Training Budget Constraints}
In this work, we explore two realistic settings corresponding to different budgets discussed in Section \ref{sec:background}. In the \textit{compute-constrained} scenario, we have less compute than data, so any data mix discards much of the available data. In the \textit{data-constrained} scenario, we have more compute than we have data, so most data mixes will require epoching over at least one of, if not all of, our datasets.

\paragraph{Compute-Constrained Experiments} Prior works on learned data mixes have focused on this setting~\cite{DoReMi, ODM}, training for 50-100B tokens. Our first set of experiments aligns with this as our largest model is trained for 100B out of 2.1T tokens. However, frontier models are increasingly trained
longer and becoming ``data-constrained"~\citep{DCScalingLaws}.

\paragraph{Data-Constrained Experiments } To identify data mixing methods applicable to frontier models, understanding the effects of epoching is essential for optimal performance~\citep{DFScalingLaws}. Since training for the full 2.1T tokens in Dolma is infeasible for a large number of baselines, we instead simulate data constraints using sub-sampling. 

Our simulation sub-samples each dataset to have $T \cdot \frac{D_{t}}{D_{s}}$ tokens where $T$ is the total tokens in the dataset, $D_{t}$ is the number of tokens we will actually train for, and $D_{s}$ is the number of tokens we are simulating constraints for. This causes the epoching behaviour in the experiment to behave as it would at the target budget. In this paper, we target a 1.6T token budget - the number of tokens seen using the OLMo V1.7 weights adjusted for the Llama tokenizer. 

\subsection{Evaluation Tasks and Metrics}

\paragraph{Per-Task Evaluation Metrics} Evaluating methods of Large Language Model pretraining with respect to downstream performance is often challenging, since discrete metrics like accuracy can be noisy at smaller scales~\citep{Emergence, EmergenceMirage}. On the other hand, negative log-likelihood (NLL) on pretraining data may not correlate with model utility for downstream tasks. 

To strike a balance between scaling predictability and correlation with downstream performance, we utilize the NLL per token \textit{on the correct answers from downstream benchmarks} as our metric across tasks. For multiple-choice tasks, we normalize the NLL by the probability assigned to all options to produce a metric that correlates with accuracy improvements but improves predictably with scale~\citep{HardToPredict, Llama3}.

We evaluate the impacts of data mixing on these metrics across benchmarks in 5 commonly-tested LLM capabilities: coding~\citep[HumanEval; MBPP]{HumanEval, MBPP}, mathematics~\citep[MATH]{MATH}, translation~\citep[FLoRes]{Flores}, reasoning~\citep[ARC]{ARCChallenge}, and general knowledge~\citep[MMLU]{MMLU}. We use the NLL over the correct answer for MBPP, HumanEval, MATH, and FLoRes and normalized NLL over the correct answer for ARC and MMLU.

\paragraph{Across-Task Evaluation Metrics}
We also report measures to ease comparison across tasks. 

\textit{Speedup} or \textit{slowdown} is the ratio of FLOPs it takes to achieve the same performance as a baseline at a specific point based on the fit scaling curve for each task. In our experiments, we provide this measure with comparison to the performance at the largest scale of $3 \times 10^{21}$. 

\textit{Mean rank} is a metric across all tasks. First each approach is ranked based on performance within each task, these ranks are then averaged across all tasks. We draw this metric from \citet{CorrelTristan} as a way to succinctly capture the relative performance of methods across several tasks.

\section{Generalization of Data Mixing Methods}
\begin{figure}[t]
\begin{center}
\includegraphics[width=1\linewidth]{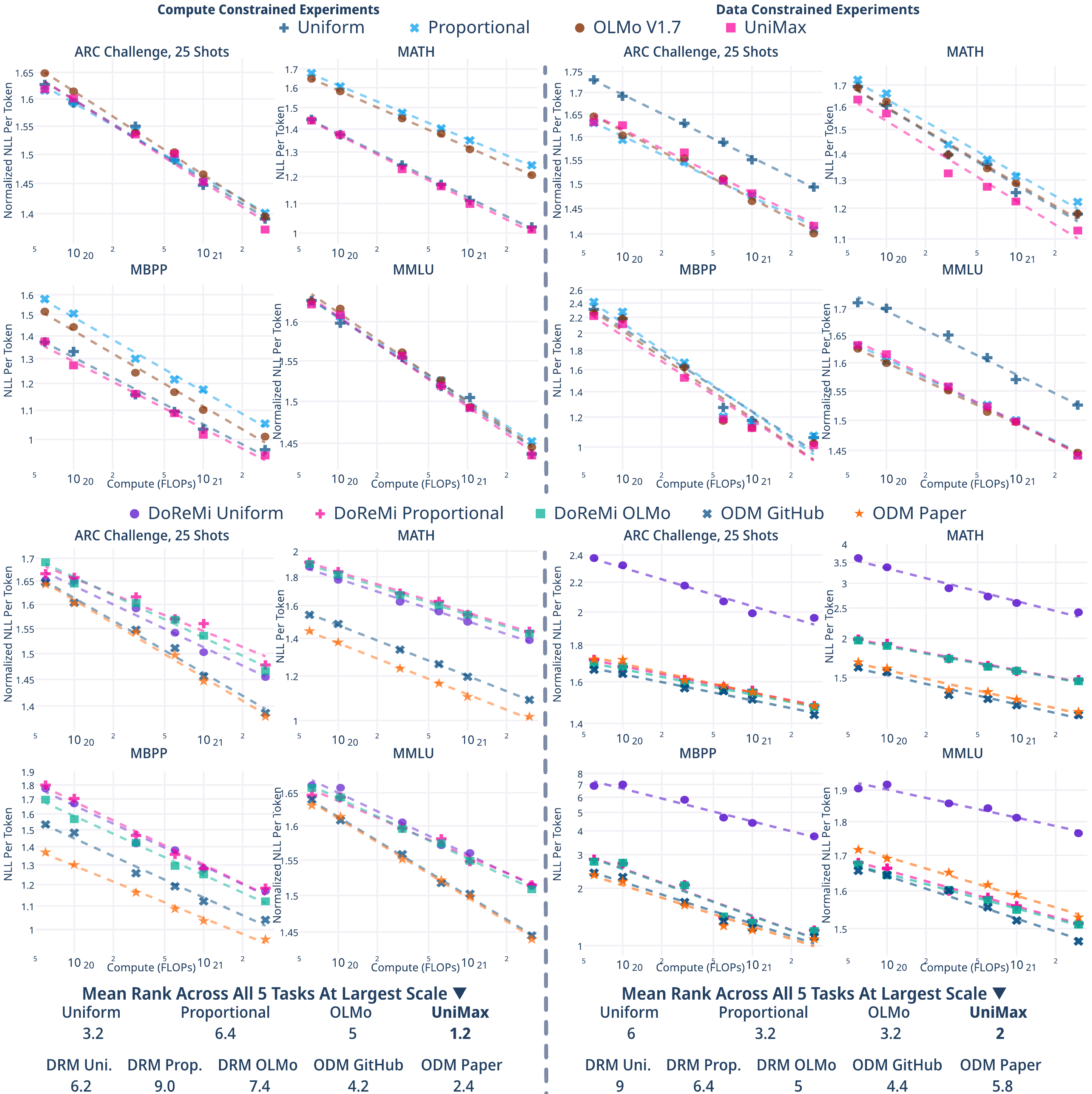}
\end{center}
\caption{Comparison of baseline data mixing methods. The model with the top average rank at $3\times10^{21}$ FLOPs is in \textbf{bold}. UniMax consistently outperforms all other baselines in both settings.}
\label{fig:baseline_compare}
\end{figure}

Model configuration, tokenizers, and shuffling are all likely to impact data mix experiments. To control for these confounders, we re-implement\footnote{For learned data mixes, we reference the following open-source code released by authors to verify our re-implementations in addition to the released papers: \href{https://github.com/alon-albalak/online-data-mixing}{ODM Github} and \href{https://github.com/sangmichaelxie/doremi/tree/main}{DoReMi GitHub}.} and re-compute data mixes for Dolma using a unified setup and consistent random seed. In \Figref{fig:baseline_compare}, we plot the comparison of all baselines.

\subsection{Baselines} \label{sec:baseline}
We compute Proportional and Uniform data mixes using the formulas from Section \ref{sec:related}. For OLMo V1.7, we use the sampling proportions shared by the authors. We re-implement the UniMax algorithm using CVXPY~\citep{CVXPY} and compute two data mixes: one for our ``compute-constrained" scenario and another for our ``data-constrained" scenario. In the 100B scenario, we use a single epoch cap while for the 1.6T token budget we use the two epoch cap used by OLMo V1.7.

We use DoReMi and Online Data Mixing (ODM) as our baselines for learned data mixes. While DoReMi uses the Proportional prior for reference models, it is unclear whether this is the optimal prior for Dolma. To account for this, we train three DoReMi variants with Uniform, Proportional, and OLMo V1.7 priors. Reference and proxy models are trained for $6 \times 10^{19}$ FLOPs and are trained separately for compute- and data-constrained experiments. 

In the ODM paper, given reward $\hat{R}$, the weights $\vw_t$ at step $t$ are computed as $(1-K\mathcal{E}_{t})\sigma(\mathcal{E}_{t-1}\hat{R})+\mathcal{E}_{t}$, where $\sigma$ denotes the softmax. We call this variant ``ODM Paper". However, $\mathcal{E}_{t-1}\rightarrow0$ as $t \rightarrow \infty$ which causes the softmax to become uniform independent of the rewards across datasets. We noted that the open-source release of ODM removes $\mathcal{E}_{t-1}$ from the softmax, and confirm with the first author that the reported experiments used this code. We call this variant ``ODM Github".

\subsection{Results}
\paragraph{Generalization to data-constraints} Our results emphasize the importance of simulating data constraints that occur in frontier training. Methods behave dramatically differently at different budgets. Data mixes which are close to uniform perform well in compute-constrained settings and perform poorly in data-constrained settings, while the opposite is true for near proportional data mixes. Reliable data mixing methods should perform well in both settings.

\paragraph{The Unreasonable Effectiveness of Uniform Utility} Our second finding is that UniMax outperforms other methods in \textit{both} settings. Given the simplicity of UniMax it may be surprising that it outperforms learned and manual baselines. However, the effectiveness of assuming uniform utility is consistent with results in portfolio optimization~\citep{OptimalvsDiversity}.

The superior performance of UniMax suggests that maintaining data diversity and scale is the major driver of performance, particularly as training runs become data-constrained. Furthermore, it makes no assumptions of downstream tasks and can be computed at near zero cost. Due to these results we compare primarily to UniMax throughout the rest of this work for clarity, with full results across all baselines and experiments reported in \ref{app:full-exp-res}.

\section{Estimated Data Utility Optimization}
\begin{figure}[t]
\begin{center}
\includegraphics[width=1\linewidth]{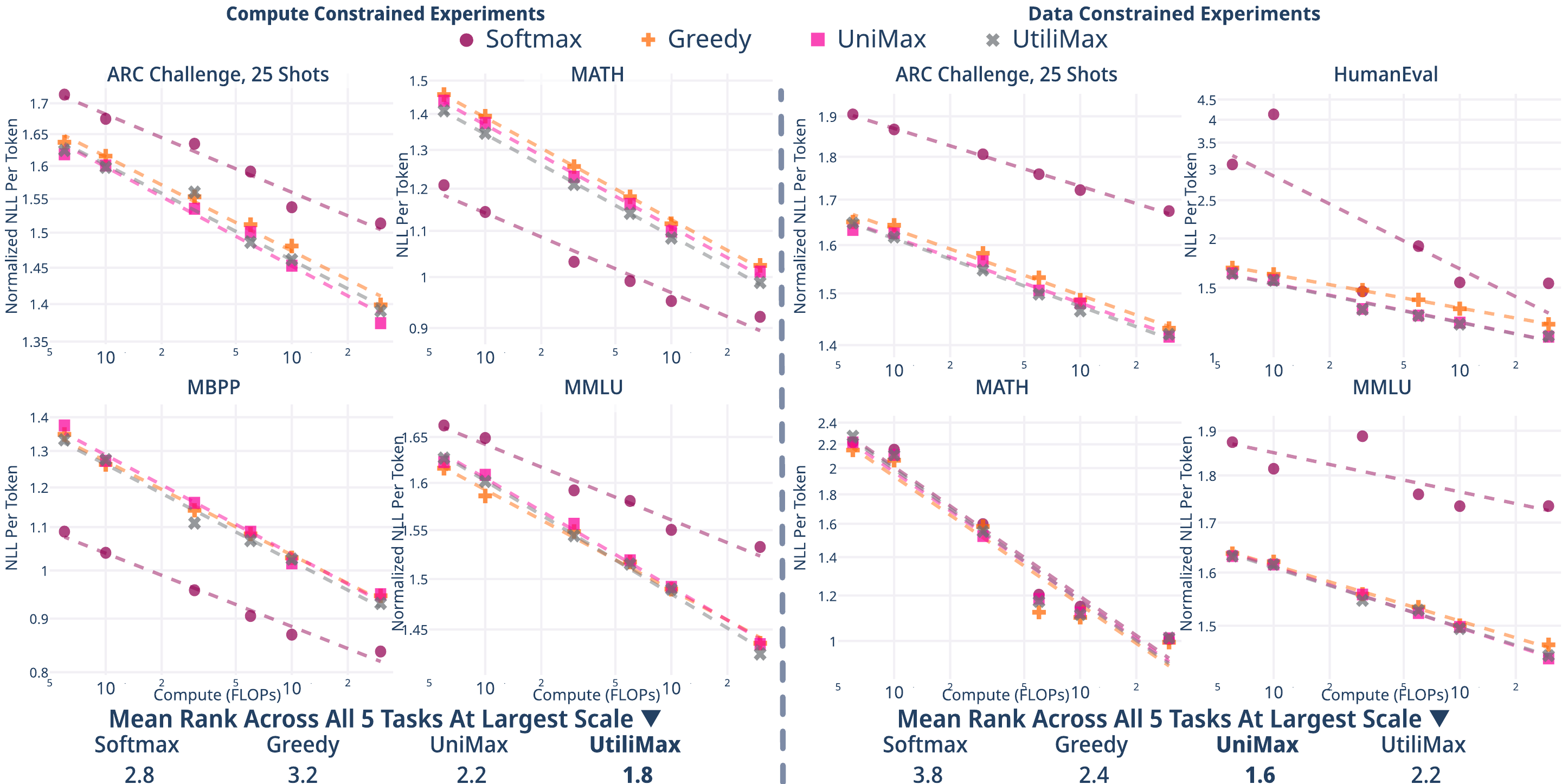}
\end{center}
\caption{Comparison of utility optimization methods. The model with the top average rank at $3\times10^{21}$ FLOPs is in \textbf{bold}. UtiliMax outperforms alternative optimization procedures in both settings.}
\label{fig:optimize_compare}
\end{figure}

Despite the effectiveness of UniMax, it is reasonable to believe that data sources have non-uniform utility for downstream tasks. The challenge is in estimating these utilities accurately and robustly handling estimation errors. In \Figref{fig:optimize_compare}, we compare our proposed method, UtiliMax, using performance on validation sets as an estimate of utility.

\subsection{Isolated Data Ablations}
 As an intrinsic measure of utility, we train proxy models for $6 \times 10^{19}$ FLOPs and evaluate on downstream tasks. We use the released held-out validation splits for this, except for MBPP where we use HumanEval as held-out validation data. We estimate utility for each Dolma dataset, treating each perplexity bucket of CommonCrawl data separately, resulting in seventeen proxy models. 

Given a set of $\sD$ datasets and $\sT$ downstream evaluations, this gives us a metric matrix $\mM \in \mathbb{R}^{|\sD|\times |\sT|}$. Since NLL metrics for each task vary significantly in scale we  normalize the range of values for each task to a normal distribution with mean $0.5$ and range $[0, 1]$ creating a utility matrix $\mU$. 

\vspace{-1em}

\subsection{UtiliMax Optimization Methodology}
In \eqref{eq:abstract}, we formulate data mix optimization as maximizing risk-adjusted utility in abstract. Here, we describe the specific problem that we solve using the Splitting Conic Solver~\citep{SCS1, SCS2} through CVXPY~\citep{CVXPY}.

UtiliMax maximizes utility by minimizing the $L_2$ distance between the expected utility vector $\vw^\intercal\mU$ of our data mix across tasks and a theoretical optimal data mix which has a utility of 1 for all tasks. 

In this work, we estimate risk associated by assuming that increasing allocation to a single dataset linearly corresponds to the number of alternative datasets. Therefore we set the risk term to $|\sD|\vw^\intercal\vw$. This could also be interpreted in two ways: (1) as maximizing utility with a specific $L_2$ regularization or (2) as interpolating between the utility maximizing solution and UniMax dependent on $|\sD|$ .  

Finally, following UniMax, we set an epoching cap $C$ on each dataset. With all of this established, UtiliMax is formulated concretely as follows:

\begin{equation}
    \argmax_{\vw} ||\vw^{\intercal}\mU - \pmb{1}||_2 + |\sD|\vw^\intercal\vw \quad \text{subject to} \quad \pmb{1}^\intercal\vw = 1, \texttt{ min}(\vw) > 0, \texttt{ }\frac{B_T\cdot\vw}{\vt} \leq C
    \label{eq:UtiliMax}
\end{equation}

We compare this approach to three alternative optimization procedures. First, simply projecting our NLL results to a valid distribution using the softmax function $\sigma(\pmb{1}^\intercal\mM)$. 
Second, using utility optimization without risk-adjustment, removing the $|\sD|\vw^\intercal\vw$ term in \eqref{eq:UtiliMax}. 
Finally, considering only risk-minimization, which is equivalent to UniMax.

\subsection{Results}

Our experimental results shows that diversity, in addition to utility, is integral to effective data mixing. Using either the softmax or greedy optimization leads to poor results in at least one setting, despite using natural estimates of dataset utility at significant computational cost. Only with the added risk-adjustment in UtiliMax does the information from ablations lead to consistent performance and improve over UniMax in either setting.

 However, using ablations to estimate utility has major shortcomings compared to assuming uniform utility. First, the quality of ablation utility estimates depends on the validation set of the benchmark. Increasingly, high-quality LLM benchmarks, such as GPQA~\citep{GPQA} or HumanEval~\citep{HumanEval}, have either very small or non-existent validation sets. Furthermore, even when large high-quality validation sets do exist, running thes ablations can be prohibitively expensive.

\begin{figure}[t]
\begin{center}
\includegraphics[width=1\linewidth]{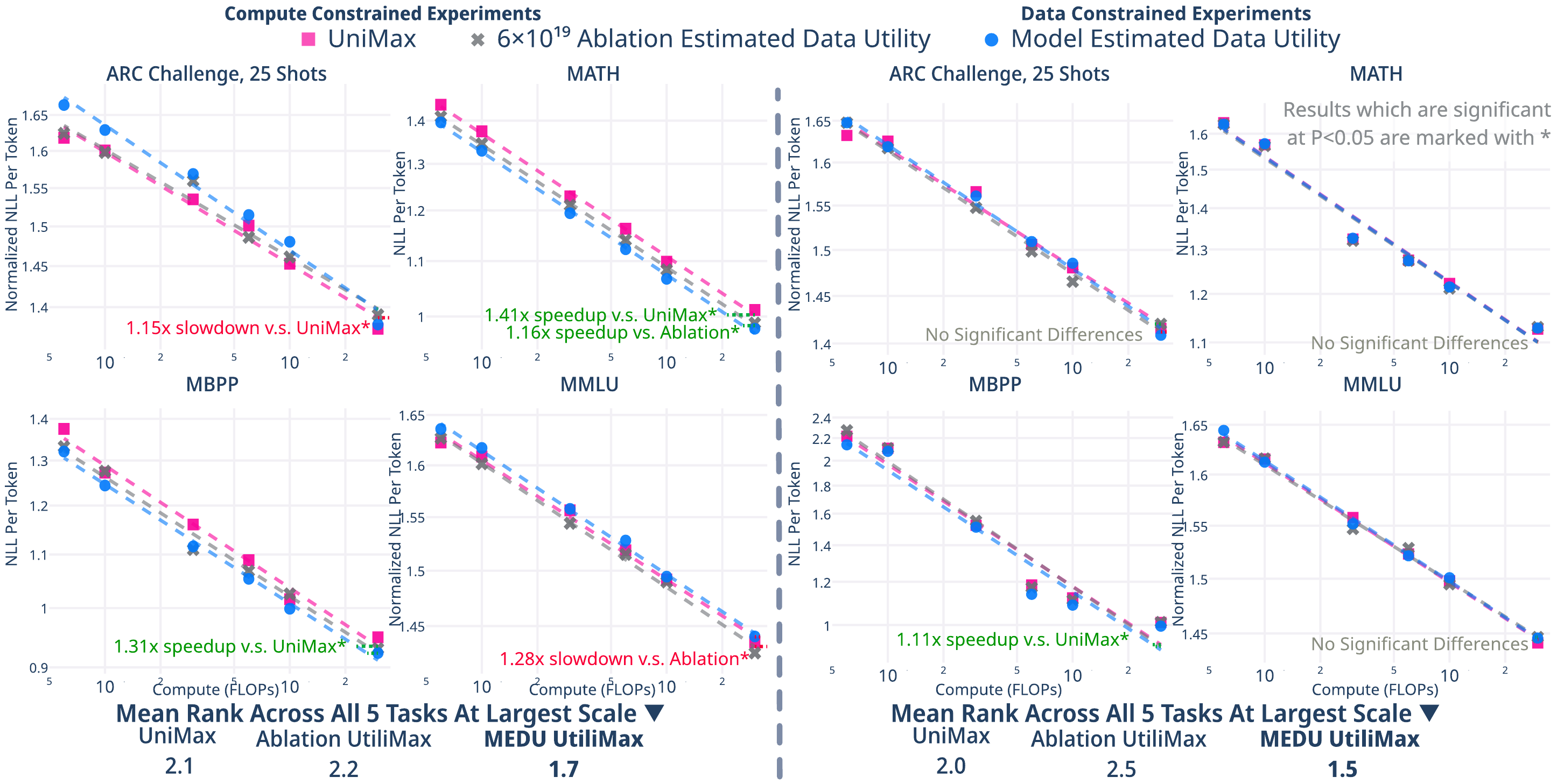}
\end{center}
\caption{Scaling curves comparing MEDU, Ablation Estimates, and UniMax. The model with the best mean rank at $3\times 10^{21}$ FLOPs is marked in \textbf{bold}. Speedup indicates cases where using MEDU improves over Ablation-Based Utility, while slowdowns indicates the opposite.}
\label{fig:MEDU_compare}
\end{figure}

\section{Reducing Cost of Utility Estimation With LLMs}  

In order to address these shortcomings, we propose a method to use existing LLMs to estimate data utility at a vastly reduced cost, similar to the model-based quality filtering described in \citet{Llama3}. Beyond reducing costs by removing the need for ablations, model-based utility estimates have the advantage of being able to generalize specific validation data into a more general description of desirable data formats and domains. In \Figref{fig:MEDU_compare}, we compare \textbf{M}odel \textbf{E}stimated \textbf{Data} \textbf{U}tility (\textbf{MEDU}) to ablation-based and uniform utility estimates.

\subsection{Model Estimated Data Utility Methodology}
\begin{wrapfigure}{H}{0.45\textwidth}
    \begin{minipage}{0.45\textwidth}
    \vspace{-1em}
    \includegraphics[width=1\linewidth]{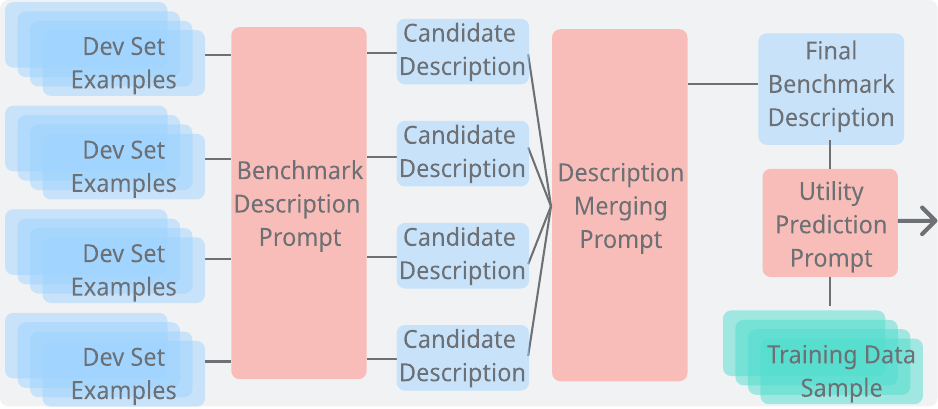}
    \caption{A high-level overview of MEDU which describes benchmarks and classifies document utility based on this description.}
    \label{fig:medu-overview}
    \end{minipage}
  \end{wrapfigure}

Our method prioritizes the following requirements. First, it must be end-to-end automated such that there is minimal prompt engineering required to incorporate new benchmarks. Secondly, it must provide estimates that are effective for UtiliMax at a fraction of the cost of ablations.

To achieve our first goal, we prompt an LLM to describe a benchmark based on the examples from the development set, then to describe the skills and knowledge required, and finally to describe documents that are likely to contain this content. In \Figref{fig:medu-overview}, we refer to this as "benchmark description".

However, many LLMs cannot fit more than a few examples at a time in their context window which could lead to sampling bias in descriptions. Therefore, we utilize hierarchical merging as proposed in ~\citet{BoookScore}. First, we generate many descriptions based on separate batches of examples. Then, we prompt the LLM to synthesize a new description from pairs of existing descriptions until only a single description remains. In \Figref{fig:medu-overview}, we describe this as "description merging".

To achieve our second goal, we use the generated description to prompt an LLM to classify which of the following utility classes best describes an individual training documents utility for the benchmark: Great, Good, Okay, Poor, or Useless. For long documents, we take a random chunk of the document using the sampling algorithm from \citet[A.1.2]{Gopher}. We map these classifications to numerical values 1, 0.75, 0.5, 0.25, 0 respectively. We call this "utility prediction" in \Figref{fig:medu-overview}. 

We utilize Llama 3 70B as the LLM for MEDU and classify relevance for a fixed random sample of documents. In practice, we find 256 documents from each corpus is effective\footnote{In \ref{app:variance_medu}, we study the variance of MEDU and sensitivity to random sampling and to LLM choice with Llama 8B, 405B, Claude Sonnet 3.5, and GPT-4o}. At the upper-bound, where each classification uses the full context length, this process requires 2.1 Million tokens. This reduces computation from $6\times 10^{19}$ FLOPs to $3\times 10^{17}$ FLOPs for MEDU, a 200x reduction.

\subsection{Results}
\begin{table*}
    \captionof{table}{Mean rank across all methods and all evaluation tasks at our largest compute scale ($3\times 10^{21}$) and averaged across all compute scales. Best rank in \textbf{bold} and the second best rank marked with *.}
    \label{tab:all_ranks}
    \centering
    \resizebox{0.8\linewidth}{!}{%
    \centering
\begin{tabular}{lcccc}
\toprule

\multirow{2}{*}{Mixing Method} & \multicolumn{2}{c}{Mean $3\times10^{21}$ Rank} & \multicolumn{2}{c}{Mean Rank Across All Scales} \\
\cmidrule(lr){2-3} \cmidrule(lr){4-5}
                                    & Compute Constrained     & Data Constrained     & Compute Constrained      & Data Constrained     \\ 
\cmidrule(r){1-1} \cmidrule(lr){2-2} \cmidrule(lr){3-3} \cmidrule(lr){4-4} \cmidrule(lr){5-5}
Uniform                    & 6.5                                     & 10.5                                 & 5.93                                    & 10.15                                \\
Proportional               & 10.4                                    & 5.6                                  & 8.88                                    & 5.88                                 \\
OLMo                       & 9.5                                     & 5.4                                  & 9.03                                    & 4.92                                 \\
UniMax                              & 3.8*                                    & 3.7*                                 & 4.68*                                   & 4.32*                                \\ \cmidrule(r){1-1}
DoReMi Uniform                      & 10.3                                    & 15.                                  & 10.62                                   & 14.88                                \\
DoReMi Proportional                 & 13.7                                    & 10.2                                 & 13.43                                   & 9.65                                 \\
DoReMi OLMo                         & 12.2                                    & 9.1                                  & 12.62                                   & 8.98                                 \\ \cmidrule(r){1-1}
ODM Github                          & 8.4                                     & 8.5                                  & 8.7                                     & 7.88                                 \\
ODM Paper                           & 5.1                                     & 10.                                  & 5.7                                     & 10.03                                \\ \cmidrule(r){1-1}
Ablation Softmax                    & 4.                                      & 10.8                                 & 4.8                                     & 11.95                                \\
MEDU Softmax                        & 9.2                                     & 12.2                                 & 9.03                                    & 12.22                                \\ \cmidrule(r){1-1}
Ablation Greedy                     & 8.2                                     & 6.7                                  & 7.75                                    & 5.92                                 \\
MEDU Greedy                         & 11.2                                    & 4.4                                  & 9.12                                    & 5                                   \\ \cmidrule(r){1-1}
Ablation UtiliMax                   & 4.1                                     & 5.                                   & \textbf{4.37}                           & 4.35                                 \\
MEDU UtiliMax                       & \textbf{3.4}                            & \textbf{2.9}                         & 5.33                                    & \textbf{3.87}                        \\ \hline

\end{tabular}
    }
\end{table*}

Our comparison of UniMax, Ablation-Based UtiliMax, and MEDU-Based UtilMax in \Figref{fig:MEDU_compare} shows that MEDU does not significantly change results: UtiliMax outperforms UniMax using \textit{either} MEDU or ablations in the compute-constrained setting\footnote{In \ref{app:relev_opt_ablate}, we repeat the UtiliMax ablations from \Figref{fig:optimize_compare} and reconfirm each term is essential.}.  Furthermore, in data-constrained settings, using MEDU \textit{improves} results  on average at the largest scale compared to running ablations though the differences between all methods is often insignificant in this setting. 

In \Tabref{tab:all_ranks}, we show that across \textit{all} methods both UtiliMax approaches achieve the best mean ranks, both in the largest scales and across all scales. Given that MEDU is much cheaper to compute, combining MEDU with UtiliMax is a pareto-optimal data mixing approach.

\begin{figure}[t]
    \centering
    \includegraphics[width=0.6\linewidth]{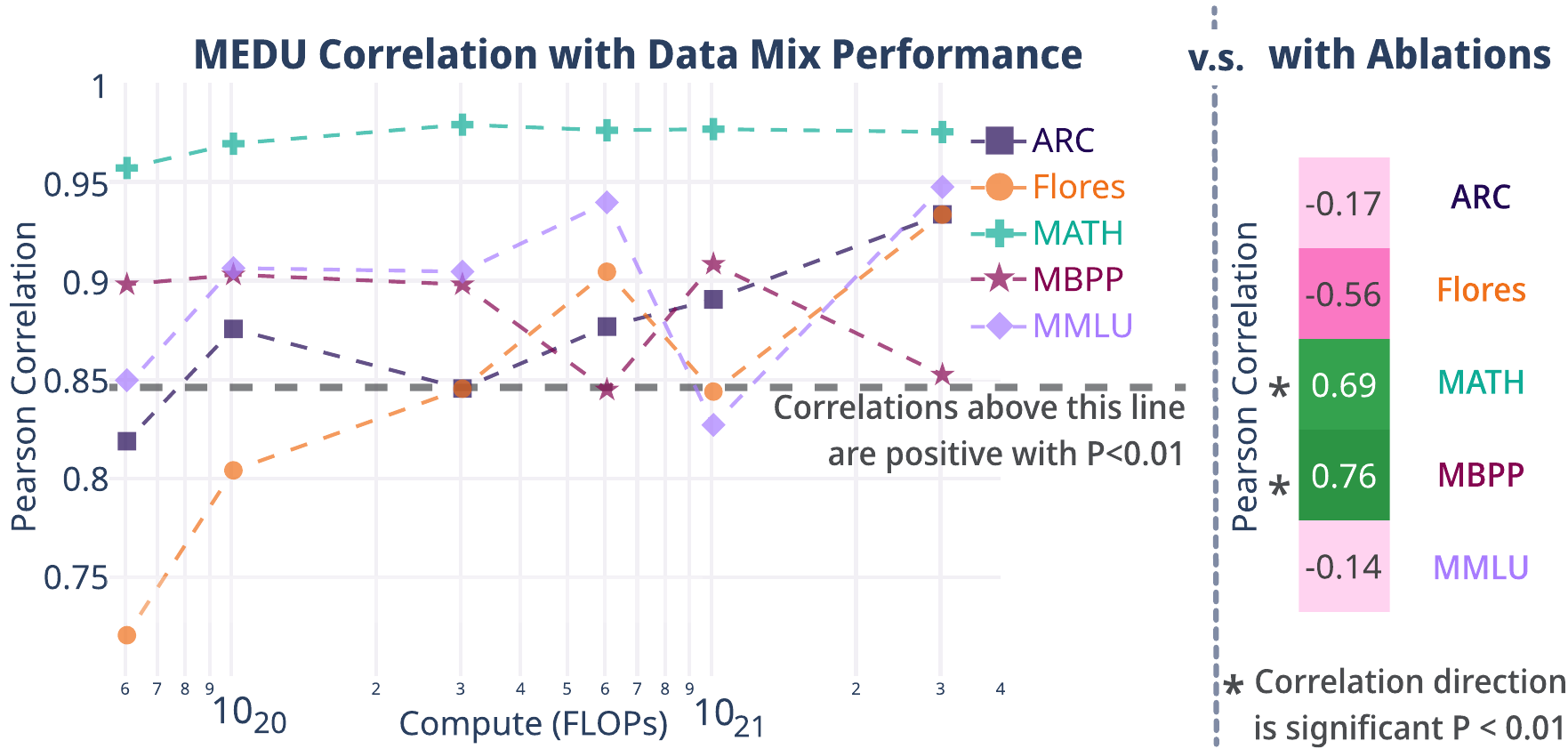}
    \caption{(left) Correlation between true performance and MEDU for all static baseline data mixes from Section \ref{sec:baseline}. (right) Correlation between MEDU and ablation estimates. Correlation is positive for MATH and MBPP, with no other significant correlation in either direction $(P>0.05)$.}
    \label{fig:correl_compare}
\end{figure}
In \Figref{fig:correl_compare}, we assess MEDU as a proxy for ablation results intrinsically through correlations. While the Pearson correlation between MEDU and individual dataset ablations are inconclusive, with correlations from only two out of five tasks excluding the null hypothesis\footnote{In \ref{app:correl_extend}, we show that this is likely driven by MEDU making comparatively sparse utility estimations.}, correlations between MEDU and the ground-truth performance of a data mix are consistently positive.

We use our baseline methods from Section \ref{sec:baseline} as the sample population and compute the Pearson Correlation between the weighted average of MEDU scores based on the data mix weights $\vw$
 and the actual results. MEDU has consistently strong positive correlation (22/30 $P<0.01$, 30/30 $P<0.05$) with ground truth performance in these full mix ablations.

\section{Conclusion}
We highlight four takewaways from this work, both broadly for experiments on data mixing methods and for the methods we propose: UtiliMax and MEDU.

\begin{itemize}
    \item \textit{Effective data mixing requires balancing utility, diversity, and scale.} While token-heuristic methods focus on diversity and scale, learned data mixing approaches focus primarily on data utility. UtiliMax succeeds largely because the optimization procedure it leverages considers all three factors, allowing to produce effective data mixes in a pareto-optimal fashion when combined with MEDU. Future work should consider improving the measures of each of these factors, such as $\Sigma$ derived from multiple proxy models as in~\citet{CorrelTristan} with principled covariance estimators~\citep{CovarianceShrinkage}.As our understanding of the utility of data increases, we see the potential of UtiliMax to serve as a principled approach for converting any utility estimate into a data mix to help train better LLMs faster.

    \item \textit{Scalable text analysis from LLMs can improve LLMs themselves.} Across disciplines~\citep{AlpacaFarm, LLMSForCSS, LLMsForPsych, LLMsForLaw}, LLM-based data analysis is becoming a common approach to scale ``qualitative" data analysis into quantifiable metrics. Although these likely contain measurement error and variance~\citep{LLMShortcomings}, as with all metrics, we show that they can be combined with principled approaches which account for this to improve the models themselves.

    \item \textit{Data mixing experiments must consider intended token-budget.} At small token budgets, some data mixes, such as uniform sampling, are very effective. However, only some of these approaches generalize to data mixes at higher budgets. Similar is true in reverse, with some methods which are weak at small-scales performing well at larger budgets. For data mixing methods to deliver predictable value, they must be tested under varied constraints.
    
    \item \textit{An extremely simple baseline, UniMax, outperforms subsequent data mixing work.} This result is consistent across settings. UniMax even performs on par with UtiliMax for multiple benchmarks which makes it a good baseline comparison candidate for those introducing new methods.

\end{itemize}

\bibliography{iclr2025_conference}
\bibliographystyle{iclr2025_conference}
\newpage
\appendix
\section{Appendix}
\ificlrfinal
\subsection{Contributions}
 Will and Todor were the project leads for this work. Todor scoped the project, conceptualized MEDU, and oversaw the project from start to finish. Will made MEDU concrete, conceptualized UtiliMax, and ran all experiments presented in this work. Bhargavi, Mike, and Frank all contributed core ideas resulting in the UtiliMax algorithm. Bhargavi debugged and refined early UtiliMax implementations. Frank led efforts which identified and defined the evaluation methodology for scaling experiments. Punit implemented the simulated epoching sub-sampler to enable data-constrained experiments. All authors contributed to writing and refining the paper.

\subsection{Acknowledgements}
This work was made possible by the collective expertise of the entire Llama team at Meta AI, especially the Pretraining Data team. Additionally, experiments would not have been possible without the Meta ML infrastructure teams which made running large-scale training runs and Llama inference jobs a smooth process.

We are grateful to Sang Michael Xie and Alon Albalak for their helpful discussion of details on their related works. Furthermore, we would like to thank Kushal Tirumala, Niladri Chatterji, Tristan Thrush, and Mirac Suzgun for conversations and comments which improved this work.

\fi

\subsection{Optimizer Ablations on Relevance Scores}\label{app:relev_opt_ablate}
\begin{figure}[H]
\begin{center}
\includegraphics[width=0.9\linewidth]{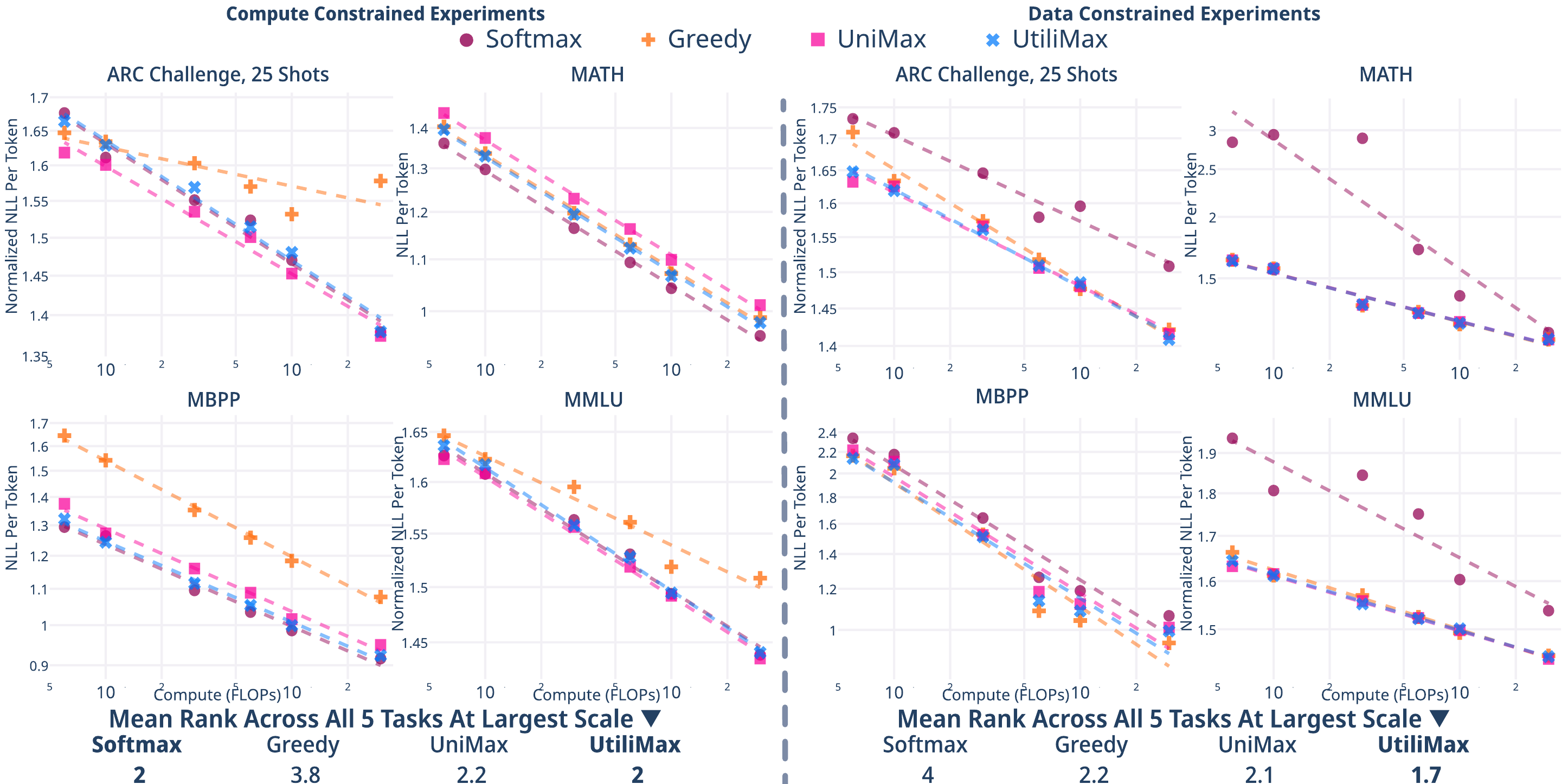}
\end{center}
\caption{Scaling curves across utility optimization methods for MEDU. The model with the top average rank for ARC, MBPP, MATH, MMLU, and FloRes at $3\times10^{21}$ FLOPs is in \textbf{bold}. UniMax outperforms greedy optimization when compute-constrained, but UtiliMax performs best in both settings, re-validating the results in \Figref{fig:optimize_compare}}
\end{figure}

\subsection{MEDU Variance and Sensitivity Analysis}\label{app:variance_medu}
\subsubsection{Variance Induced by Random Sampling}
\begin{figure}[H]
    \centering
    \includegraphics[width=1\linewidth]{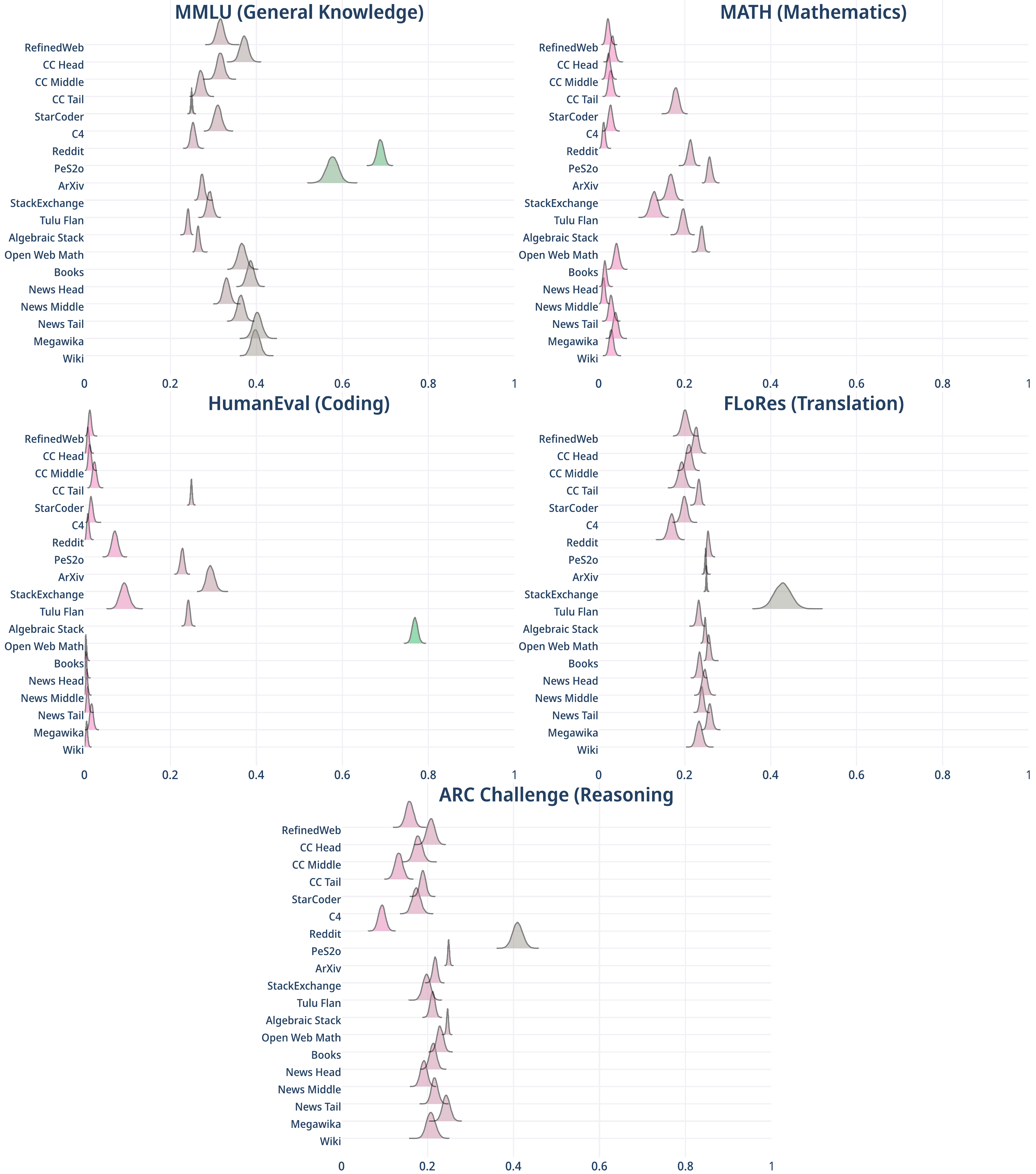}
    \caption{Distribution of MEDU Scores for a sample of 256 documents using the Llama 70B model. To compute these distributions, we run MEDU on a sample of 1024 documents, then recompute the mean 10,000 times using bootstrap sampling. At 256 examples, the distributions are tight enough that larger sample size would minimally impact the data mixes produced by UtiliMax.}
    \label{fig:joy-var-}
\end{figure}

\subsubsection{Variance Across Model Choices}\label{app:model_opt_ablate}
\begin{figure}[H]
\begin{center}
    \begin{minipage}{0.53\textwidth}
\includegraphics[width=1\linewidth]{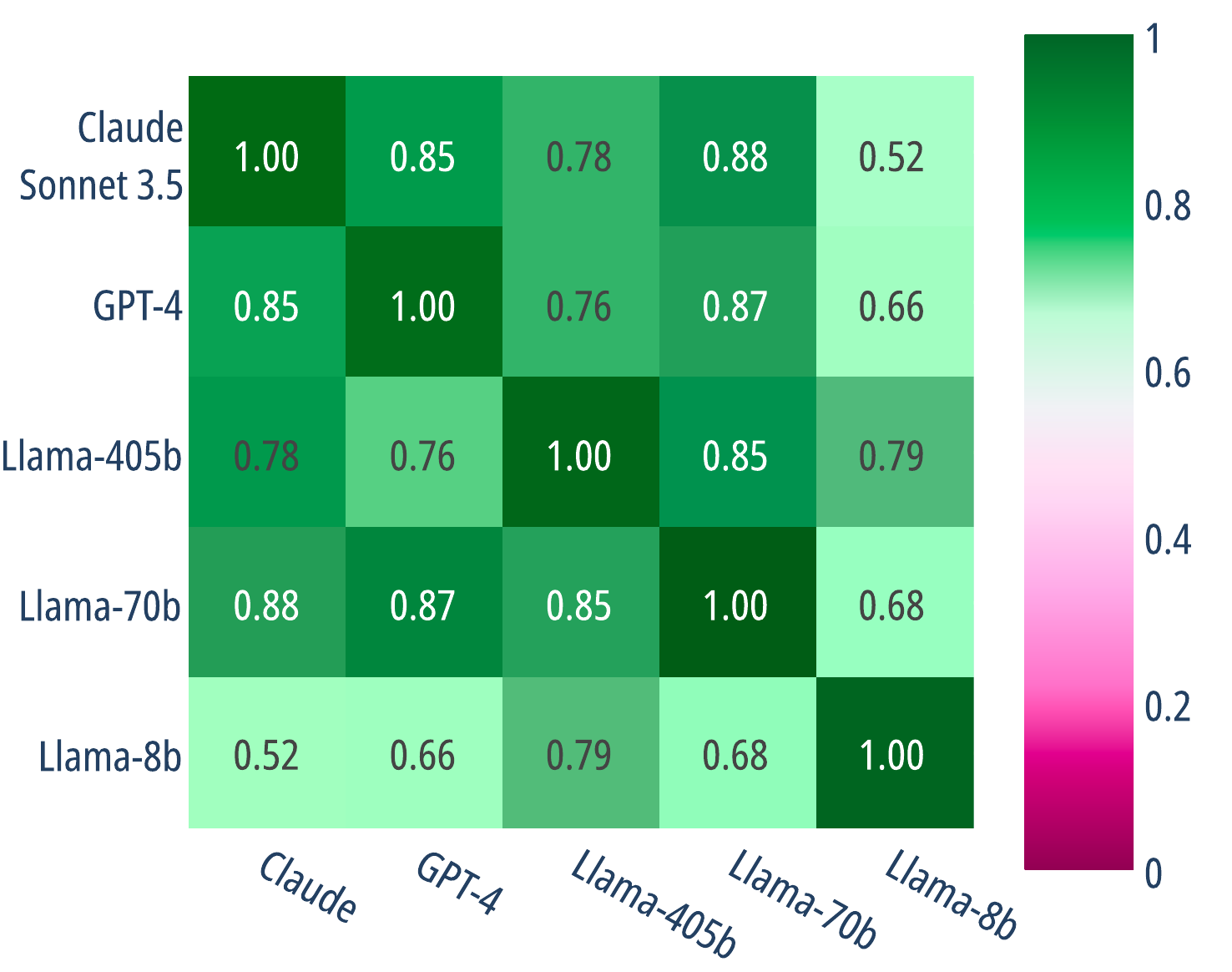}
\caption{Pearson correlations of model-estimated data utility using different language models and different samples of data. In general, except for the 8B Llama model, we see strong correlation ($>0.75$) across tested models.}
\label{fig:correl_btw_models}
\end{minipage}
\hfill
    \begin{minipage}{0.45\textwidth}
\includegraphics[width=1\linewidth]{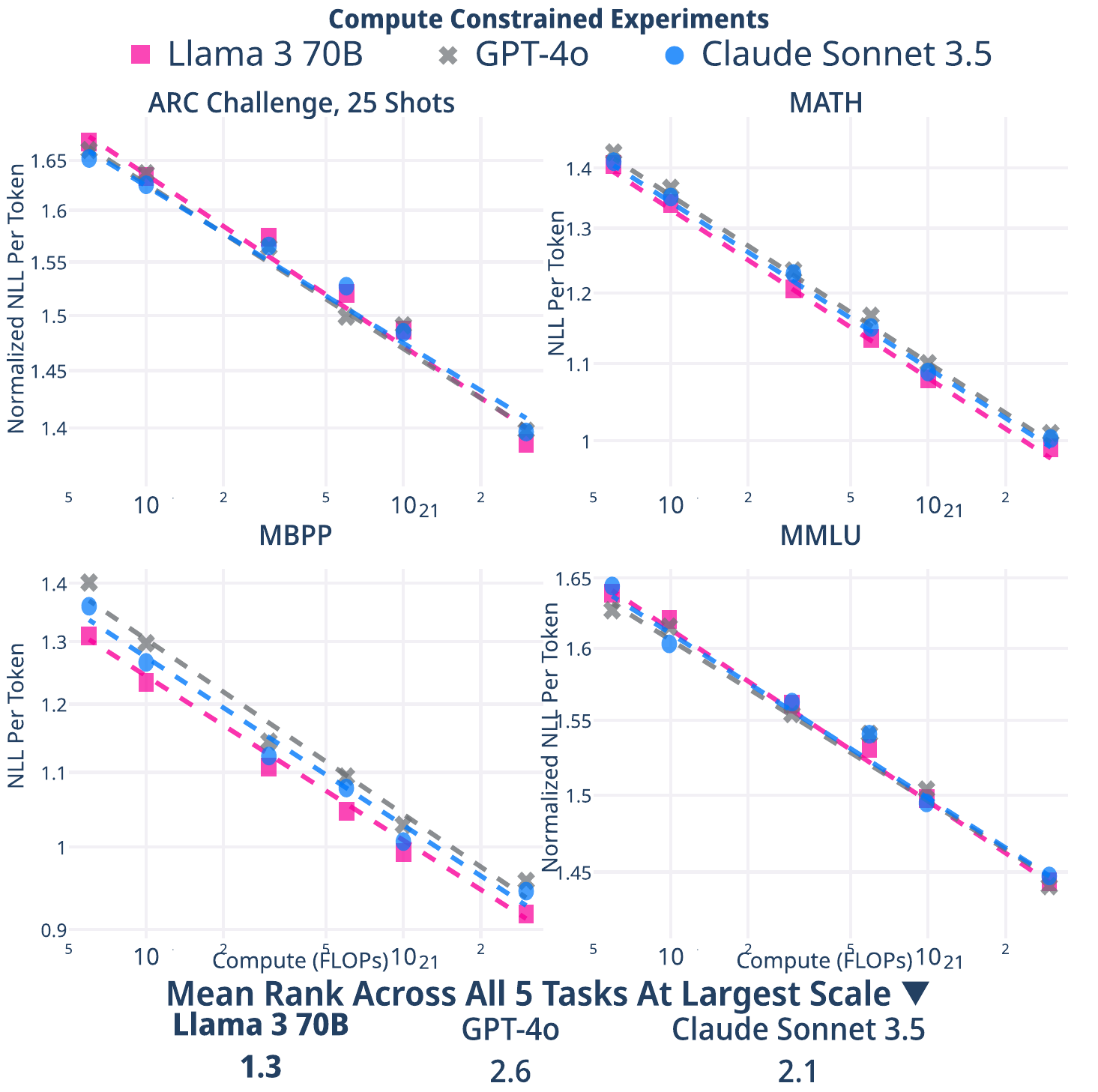}
\caption{Scaling curves across different classifier models methods for MEDU. Results are extremely similar across models suggesting that there is not a significant amount of bias injected by a specific LLM.}
\label{fig:perf_across_models}
    \end{minipage}
\end{center}
\end{figure}
For the core experiments in the paper, we utilize the Llama 3.1 70B model. This scale was selected as it offers strong performance, but can still be run on a single node for inference. Furthermore, since it is open-access it makes it easier to reproduce our results, while API models are subject to break reproducibility with version changes.

However, in order to assess how sensitive MEDU may be to the selection of an underlying language model we did further analysis on several different language models. First, we assess the effects of scale comparing Llama 3.1 8B, 70B, and 405B. Then, we assess the variance across frontier model families by comparing Llama, GPT, and Claude models.

For each model, we use the same prompts provided in Appendix \ref{app:propmpts}. This means that each model is used for the entire process, including generating benchmark descriptions which are then used for utility classification.

In \Figref{fig:correl_btw_models}, we visualize the Pearson correlations between utility-scores estimated using each model. Overall, the correlation is strong ($>0.75$) for all models, except for Llama 3 8b. This suggests that, even with different models, MEDU tends to capture similar signal from the underlying data. While this signal may not be optimal, this consistency is important as it suggests that methods built on top of model-estimated data utility are unlikely to see significant shifts due to model selection, at least within the current generation of frontier models.

To add further evidence to this, in \Figref{fig:perf_across_models}, we run full experiments of MEDU UtiliMax with Claude Sonnet 3.5 and GPT-4o. Since our experiments showed that data mixing methods have the largest impact in compute constrained settings, we run this set of experiments only in that setting. Overall, we see that the performance of MEDU UtiliMax is robust to model selection, with Llama 3 70B performing the best by a small margin overall but very little variance in downstream performance in any individual task.

\vspace{-0.1in}
\subsection{Comparing MEDU and Ablation-Based Utility Estimates}\label{app:correl_extend}
\vspace{-0.2in}
\begin{figure}[H]
    \centering
    \includegraphics[width=0.7\linewidth]{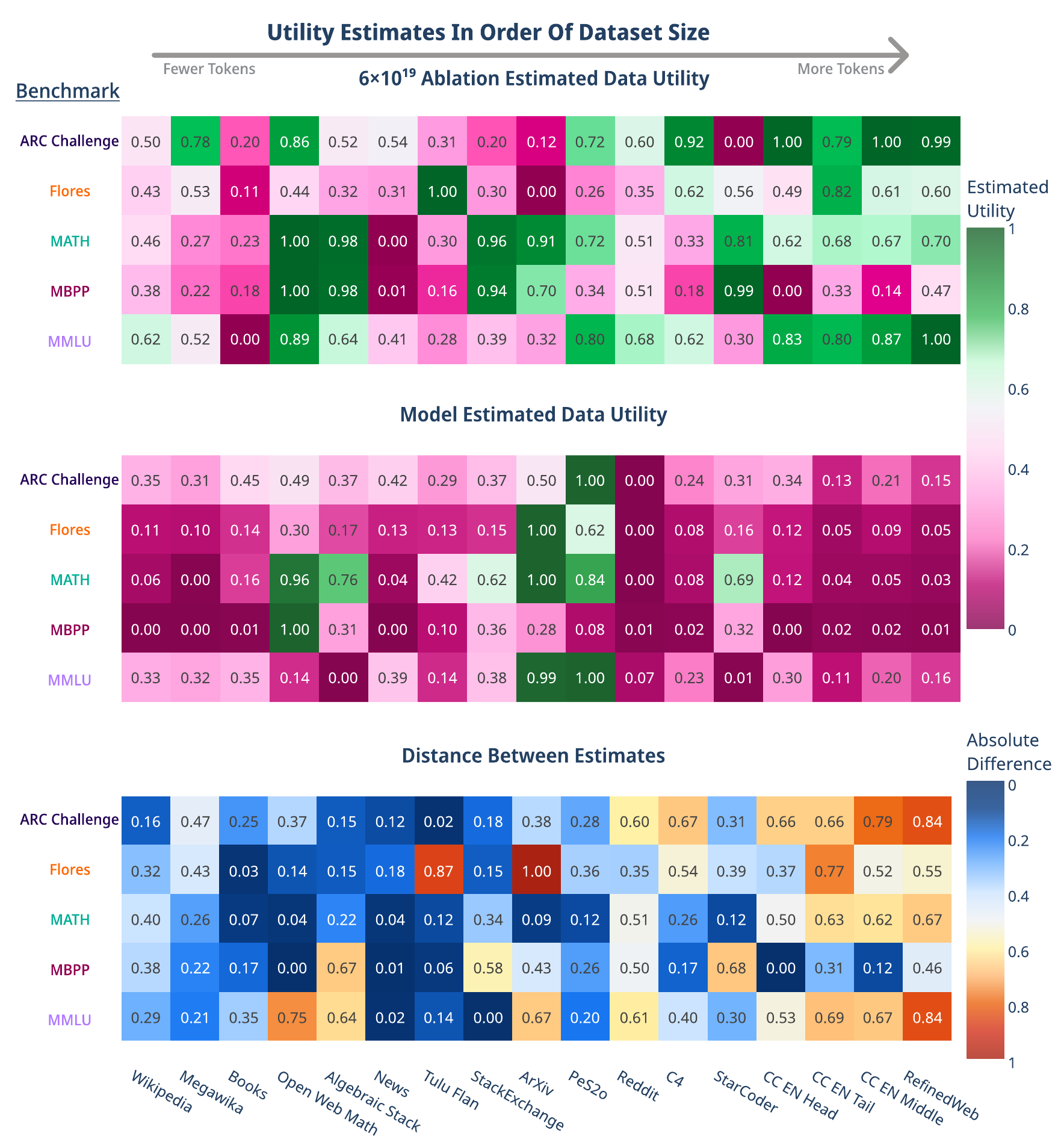}
    \caption{Heatmaps of utility scores across the two methods, with increasingly large datasets placed to the right. While ablations provide a set of estimates closer to a normal distribution, MEDU assigns high utility to just a few outliers for each task, with the rest of values skewing below the mean. MEDU also seems to systematically give lower scores to large web-corpora.}
    \label{fig:enter-label}
\end{figure}

\subsection{Prompts}\label{app:propmpts}
\subsubsection{Benchmark Description}
\begin{lstlisting}
f"""
{corpus}
Help me decide the types of training data to look for to train a language model for an evaluation with data similar to the above. 
You should keep the description brief and it is okay to generalize or abstract specific details to do so. 
Give your answer in three sections, first write what type of test this might be from, then write out the languages, skills and knowledge the language model would need, and finally write a description of the ideal training data for the evaluation.
"""
\end{lstlisting}

\subsubsection{Description Merging}
\begin{lstlisting}
f"""
<BEGIN CORPUS DESCRIPTION A>
{description_a}
<END CORPUS DESCRIPTION A>
<BEGIN CORPUS DESCRIPTION B>
{description_b}
<END CORPUS DESCRIPTION B>
{comparison}
The above analyses were written about a NLP evaluation used for Large Language Models by two different people based on equally sized random samples of examples from the evaluation. 
Help me synthesize them into a more complete analyses based on both of them. You should keep the description brief and it is okay to generalize or abstract specific details to do so. 
Give your answer in three sections, first write what type of test this might be from, then write out the languages, skills and knowledge the language model would need, and finally write a description of the ideal training data for the evaluation.
"""
\end{lstlisting}

\subsubsection{Utility Classification}
\begin{lstlisting}
f"""
The following document is being considered as training data for a Large Language Model.

Provide a concise description of the document and an assessment of the quality of the text or code in the document.

Key Attributes to Mention
- Languages contained in the document
- The coherence of the document
- The skills the document demonstrates
- The topics the document contains facts and information about

Document{prompt_addition}: 
```
{example}
```

Based on your previous reasoning, give me a concrete decision about the utility of the document as training data for the following benchmark. If a benchmark is Multilingual, you should assume a high-degree of importance is placed on high-quality content in languages other than English.

{test_description}
        
Output your decision about the utility of the data as one of the following single words Great/Good/Okay/Poor/Useless without formatting.==
"""
\end{lstlisting}
\subsection{Data Shuffling, Sampling, and Packing}\label{app:data_loader}
Consider a data mix containing a list of datasets $D_i$ with their associated weights $w_i$ , where $i\in\{1..n\}$. We assume here that the weights $w_i$ have been normalized to 1. Additionally, let us suppose that the overall batch size is $B$ and sequence length is $S$. Therefore, in each step, we need to sample $B$ sequences, each of length $S$ tokens, from the different datasets $D_i$.

In the beginning of training, we initialize a dataset iterator for each dataset $D_i$, denoted as $Iter(D_i)$, which is responsible for packing together $S$ tokens from dataset $D_i$ into a dense sequence. A seed $seed(epoch)$ which is a function of the epoch count is specified to determine how the dataset is shuffled. $Iter(D_i)$ will sample as many complete documents as necessary from $D_i$ to fill in $S$ tokens, keeping any remaining tokens in a buffer to be used in the next iteration.

The overall batch creation process for the next step can be summarized below
\begin{algorithm}
\caption{Batch creation process}
\label{alg:example}
\begin{algorithmic}[1]
    \STATE \textbf{Input:} A list of dataset iterators $Iter(D_i)$ along with their associated weights $w_i$.
    \STATE \textbf{Output:} The next batch

    \STATE weightsVector = $[w_1, w_2 ... w_n]$

    batch = []
    \FOR {batchIndex in range(1 .. $B$)}
        \STATE \# Sample from a multinomial distribution based on weightsVector.
        \STATE chosenDatasetIndex = np.random.choice$(len(weightsVector), p=weightsVector)$
        \STATE sequence = next($Iter(D_{chosenDatasetIndex})$)
        \STATE batch.append(sequence)
    \ENDFOR

    \RETURN batch
\end{algorithmic}
\end{algorithm}

As we can see, each dataset can epoch at different times based on their own dataset weights and the number of tokens in the dataset. At the end of an epoch, we reinitialize $Iter(D_i)$ with a new $seed(epoch+1)$ for the next epoch so that we get a shuffled view of the dataset.

\newpage
\subsection{Full Experimental Results}\label{app:full-exp-res}
\begin{table}[H]
\caption{NLL Metrics for each individual run without simulated epoching. NLL Per Token is used for MATH, HumanEval, Flores, and GPQA. Normalized NLL Per Token is used for ARC Challenge and MMLU, since they are Multiple-Choice Tasks. Best results in \textbf{bold}.}
\label{tab:ne-full-res-1}
\begin{center}
\begin{tabular}{llccccc}
\toprule
\multicolumn{2}{c}{Training Run Setup}                                 & \multicolumn{5}{c}{Downstream Evaluation}   \\
\cmidrule(lr){1-2} \cmidrule(lr){3-7} 
Mixing Method  & FLOPs & ARC & Flores        & MATH          & MBPP    &    MMLU      \\ \cmidrule(lr){1-1} \cmidrule(lr){2-2} \cmidrule(lr){3-3} \cmidrule(lr){4-4} \cmidrule(lr){5-5} \cmidrule(lr){6-6} \cmidrule(lr){7-7}
\multirow{6}{*}{DoReMi OLMo}     & $6\times10^{19}$ & 1.69                  & 3.91          & 1.90          & 1.70          & 1.66          \\
                                 & $1\times10^{20}$ & 1.64                  & 3.81          & 1.82          & 1.57          & 1.64          \\
                                 & $3\times10^{20}$ & 1.60                  & 3.29          & 1.67          & 1.42          & 1.60          \\
                                 & $6\times10^{20}$ & 1.57                  & 2.91          & 1.60          & 1.29          & 1.57          \\
                                 & $1\times10^{21}$ & 1.54                  & 2.70          & 1.54          & 1.25          & 1.55          \\
                                 & $3\times10^{21}$ & 1.47                  & 2.37          & 1.43          & 1.12          & 1.51          \\ \cmidrule(l){1-2}
\multirow{6}{*}{DoReMi Prop.}    & $6\times10^{19}$ & 1.67                  & 4.00          & 1.91          & 1.80          & 1.65          \\
                                 & $1\times10^{20}$ & 1.66                  & 3.76          & 1.84          & 1.70          & 1.64          \\
                                 & $3\times10^{20}$ & 1.62                  & 3.21          & 1.68          & 1.47          & 1.60          \\
                                 & $6\times10^{20}$ & 1.57                  & 2.95          & 1.63          & 1.36          & 1.58          \\
                                 & $1\times10^{21}$ & 1.56                  & 2.74          & 1.54          & 1.28          & 1.55          \\
                                 & $3\times10^{21}$ & 1.48                  & 2.44          & 1.44          & 1.18          & 1.51          \\ \cmidrule(l){1-2}
\multirow{6}{*}{DoReMi Uniform}  & $6\times10^{19}$ & 1.65                  & 3.04          & 1.88          & 1.77          & 1.66          \\
                                 & $1\times10^{20}$ & 1.65                  & 2.93          & 1.78          & 1.67          & 1.66          \\
                                 & $3\times10^{20}$ & 1.59                  & 2.41          & 1.63          & 1.42          & 1.61          \\
                                 & $6\times10^{20}$ & 1.54                  & 2.18          & 1.56          & 1.38          & 1.57          \\
                                 & $1\times10^{21}$ & 1.50                  & 1.99          & 1.50          & 1.28          & 1.56          \\
                                 & $3\times10^{21}$ & 1.45                  & \textbf{1.73} & 1.39          & 1.17          & 1.51          \\ \cmidrule(l){1-2}
\multirow{6}{*}{ODM GitHub}      & $6\times10^{19}$ & 1.65                  & 3.86          & 1.54          & 1.53          & 1.64          \\
                                 & $1\times10^{20}$ & 1.60                  & 3.94          & 1.48          & 1.48          & 1.61          \\
                                 & $3\times10^{20}$ & 1.55                  & 3.26          & 1.34          & 1.26          & 1.56          \\
                                 & $6\times10^{20}$ & 1.51                  & 2.94          & 1.26          & 1.19          & 1.52          \\
                                 & $1\times10^{21}$ & 1.46                  & 2.59          & 1.20          & 1.12          & 1.50          \\
                                 & $3\times10^{21}$ & 1.39                  & 2.24          & 1.09          & 1.04          & 1.45          \\ \cmidrule(l){1-2}
\multirow{6}{*}{ODM Paper}       & $6\times10^{19}$ & 1.64                  & 3.86          & 1.44          & 1.37          & 1.63          \\
                                 & $1\times10^{20}$ & 1.60                  & 3.60          & 1.38          & 1.30          & 1.61          \\
                                 & $3\times10^{20}$ & 1.54                  & 3.19          & 1.24          & 1.16          & 1.55          \\
                                 & $6\times10^{20}$ & 1.50                  & 2.79          & 1.16          & 1.09          & 1.52          \\
                                 & $1\times10^{21}$ & 1.45                  & 2.50          & 1.10          & 1.04          & 1.50          \\
                                 & $3\times10^{21}$ & 1.38                  & 2.11          & 1.02          & 0.96          & 1.44          \\ \cmidrule(l){1-2}
\multirow{6}{*}{OLMo Sampling}   & $6\times10^{19}$ & 1.65                  & 4.32          & 1.65          & 1.52          & 1.63          \\
                                 & $1\times10^{20}$ & 1.61                  & 3.93          & 1.58          & 1.44          & 1.62          \\
                                 & $3\times10^{20}$ & 1.54                  & 3.23          & 1.45          & 1.24          & 1.56          \\
                                 & $6\times10^{20}$ & 1.50                  & 2.91          & 1.38          & 1.17          & 1.53          \\
                                 & $1\times10^{21}$ & 1.47                  & 2.76          & 1.31          & 1.10          & 1.49          \\
                                 & $3\times10^{21}$ & 1.40                  & 2.37          & 1.21          & 1.01          & 1.45          \\ \cmidrule(l){1-2}
\multirow{6}{*}{Greedy Ablation} & $6\times10^{19}$ & 1.64                  & 4.56          & 1.46          & 1.35          & 1.62          \\
                                 & $1\times10^{20}$ & 1.62                  & 4.30          & 1.39          & 1.26          & 1.59          \\
                                 & $3\times10^{20}$ & 1.55                  & 3.56          & 1.26          & 1.14          & 1.55          \\
                                 & $6\times10^{20}$ & 1.51                  & 3.23          & 1.18          & 1.08          & 1.52          \\
                                 & $1\times10^{21}$ & 1.48                  & 2.98          & 1.12          & 1.03          & 1.49          \\
                                 & $3\times10^{21}$ & 1.40                  & 2.65          & 1.02          & 0.95          & 1.44          \\ \cmidrule(l){1-2}
\end{tabular}
\end{center}
\end{table}

\begin{table}[t]
\caption{(Cont.) NLL Metrics for each individual run without simulated epoching. NLL Per Token is used for MATH, HumanEval, Flores, and GPQA. Normalized NLL Per Token is used for ARC Challenge and MMLU, since they are Multiple-Choice Tasks. Best results in \textbf{bold}.}
\label{tab:ne-full-res-2}
\begin{center}
\begin{tabular}{lcccccc}
\toprule
\multicolumn{2}{c}{Training Run Setup}                                 & \multicolumn{5}{c}{Downstream Evaluation}   \\
\cmidrule(lr){1-2} \cmidrule(lr){3-7} 
Mixing Method  & FLOPs & ARC & Flores          & MATH    & MBPP          & MMLU          \\ \cmidrule(lr){1-1} \cmidrule(lr){2-2} \cmidrule(lr){3-3} \cmidrule(lr){4-4} \cmidrule(lr){5-5} \cmidrule(lr){6-6} \cmidrule(lr){7-7}
\multirow{6}{*}{UtiliMax Ablation} & $6\times10^{19}$ & 1.62                  & 3.76   & 1.41          & 1.33          & 1.63          \\
                                   & $1\times10^{20}$ & 1.60                  & 3.67   & 1.35          & 1.28          & 1.60          \\
                                   & $3\times10^{20}$ & 1.56                  & 2.94   & 1.21          & 1.11          & 1.54          \\
                                   & $6\times10^{20}$ & 1.49                  & 2.70   & 1.14          & 1.07          & 1.52          \\
                                   & $1\times10^{21}$ & 1.46                  & 2.58   & 1.08          & 1.03          & 1.49          \\
                                   & $3\times10^{21}$ & 1.39                  & 2.17   & 0.99          & 0.93          & \textbf{1.43} \\
\multirow{6}{*}{Softmax Ablation}  & $6\times10^{19}$ & 1.71                  & 4.86   & 1.21          & 1.09          & 1.66          \\
                                   & $1\times10^{20}$ & 1.67                  & 4.58   & 1.14          & 1.04          & 1.65          \\
                                   & $3\times10^{20}$ & 1.63                  & 3.81   & 1.03          & 0.96          & 1.59          \\
                                   & $6\times10^{20}$ & 1.59                  & 3.49   & 0.99          & 0.91          & 1.58          \\
                                   & $1\times10^{21}$ & 1.54                  & 3.29   & 0.95          & 0.87          & 1.55          \\
                                   & $3\times10^{21}$ & 1.51                  & 2.86   & \textbf{0.92} & \textbf{0.84} & 1.53          \\ \cmidrule(l){1-2}
\multirow{6}{*}{Proportional}      & $6\times10^{19}$ & 1.62                  & 4.30   & 1.68          & 1.58          & 1.62          \\
                                   & $1\times10^{20}$ & 1.59                  & 3.96   & 1.61          & 1.51          & 1.61          \\
                                   & $3\times10^{20}$ & 1.54                  & 3.33   & 1.48          & 1.30          & 1.56          \\
                                   & $6\times10^{20}$ & 1.49                  & 2.94   & 1.40          & 1.22          & 1.53          \\
                                   & $1\times10^{21}$ & 1.45                  & 2.73   & 1.35          & 1.18          & 1.49          \\
                                   & $3\times10^{21}$ & 1.40                  & 2.39   & 1.25          & 1.05          & 1.45          \\ \cmidrule(l){1-2}
\multirow{6}{*}{Greedy MEDU}       & $6\times10^{19}$ & 1.65                  & 3.75   & 1.40          & 1.64          & 1.65          \\
                                   & $1\times10^{20}$ & 1.63                  & 3.67   & 1.34          & 1.54          & 1.62          \\
                                   & $3\times10^{20}$ & 1.60                  & 3.08   & 1.20          & 1.35          & 1.60          \\
                                   & $6\times10^{20}$ & 1.57                  & 2.80   & 1.13          & 1.26          & 1.56          \\
                                   & $1\times10^{21}$ & 1.53                  & 2.53   & 1.07          & 1.18          & 1.52          \\
                                   & $3\times10^{21}$ & 1.58                  & 2.40   & 0.99          & 1.08          & 1.51          \\ \cmidrule(l){1-2}
\multirow{6}{*}{UtiliMax MEDU}     & $6\times10^{19}$ & 1.66                  & 3.82   & 1.40          & 1.32          & 1.64          \\
                                   & $1\times10^{20}$ & 1.63                  & 3.59   & 1.33          & 1.24          & 1.62          \\
                                   & $3\times10^{20}$ & 1.57                  & 3.08   & 1.19          & 1.12          & 1.56          \\
                                   & $6\times10^{20}$ & 1.51                  & 2.47   & 1.12          & 1.05          & 1.53          \\
                                   & $1\times10^{21}$ & 1.48                  & 2.57   & 1.07          & 1.00          & 1.49          \\
                                   & $3\times10^{21}$ & 1.38                  & 2.13   & 0.98          & 0.92          & 1.44          \\ \cmidrule(l){1-2}
\multirow{6}{*}{Softmax MEDU}      & $6\times10^{19}$ & 1.68                  & 3.81   & 1.36          & 1.29          & 1.63          \\
                                   & $1\times10^{20}$ & 1.61                  & 3.70   & 1.30          & 1.26          & 1.61          \\
                                   & $3\times10^{20}$ & 1.55                  & 3.06   & 1.16          & 1.10          & 1.56          \\
                                   & $6\times10^{20}$ & 1.52                  & 2.84   & 1.09          & 1.04          & 1.53          \\
                                   & $1\times10^{21}$ & 1.47                  & 2.49   & 1.04          & 0.99          & 1.49          \\
                                   & $3\times10^{21}$ & 1.38                  & 2.28   & 0.96          & 0.92          & 1.44          \\ \cmidrule(l){1-2}
\multirow{6}{*}{UniMax}            & $6\times10^{19}$ & 1.62                  & 3.86   & 1.44          & 1.38          & 1.62          \\
                                   & $1\times10^{20}$ & 1.60                  & 3.46   & 1.37          & 1.27          & 1.61          \\
                                   & $3\times10^{20}$ & 1.54                  & 3.03   & 1.23          & 1.16          & 1.56          \\
                                   & $6\times10^{20}$ & 1.50                  & 2.61   & 1.16          & 1.09          & 1.52          \\
                                   & $1\times10^{21}$ & 1.45                  & 2.52   & 1.10          & 1.02          & 1.49          \\
                                   & $3\times10^{21}$ & \textbf{1.37}         & 2.07   & 1.01          & 0.95          & 1.44          \\ \cmidrule(l){1-2}
\multirow{6}{*}{Uniform}           & $6\times10^{19}$ & 1.63                  & 3.87   & 1.44          & 1.37          & 1.63          \\
                                   & $1\times10^{20}$ & 1.59                  & 3.72   & 1.37          & 1.33          & 1.60          \\
                                   & $3\times10^{20}$ & 1.55                  & 3.05   & 1.25          & 1.16          & 1.55          \\
                                   & $6\times10^{20}$ & 1.49                  & 2.73   & 1.17          & 1.10          & 1.52          \\
                                   & $1\times10^{21}$ & 1.45                  & 2.56   & 1.11          & 1.03          & 1.50          \\
                                   & $3\times10^{21}$ & 1.39                  & 2.19   & 1.02          & 0.97          & 1.44          \\ \bottomrule
\end{tabular}
\end{center}
\end{table}
\begin{table}[b]
\caption{NLL Metrics for each individual run with simulated epoching. NLL Per Token is used for MATH, HumanEval, Flores, and GPQA. Normalized NLL Per Token is used for ARC Challenge and MMLU, since they are Multiple-Choice Tasks. Best results in \textbf{bold}.}
\label{tab:full-res-1}
\begin{center}
\begin{tabular}{llccccc}
\toprule
\multicolumn{2}{c}{Training Run Setup}                                 & \multicolumn{5}{c}{Downstream Evaluation}   \\
\cmidrule(lr){1-2} \cmidrule(lr){3-7} 
Mixing Method  & FLOPs & ARC & Flores         & MATH    & MBPP          & MMLU          \\ \cmidrule(lr){1-1} \cmidrule(lr){2-2} \cmidrule(lr){3-3} \cmidrule(lr){4-4} \cmidrule(lr){5-5} \cmidrule(lr){6-6} \cmidrule(lr){7-7}
\multirow{6}{*}{DoReMi OLMo}         & $6 \times 10^{19}$ & 1.69                  & 4.24          & 1.97          & 2.75          & 1.67          \\
                                     & $1 \times 10^{20}$ & 1.68                  & 3.91          & 1.89          & 2.72          & 1.64          \\
                                     & $3 \times 10^{20}$ & 1.59                  & 3.38          & 1.72          & 2.07          & 1.60          \\
                                     & $6 \times 10^{20}$ & 1.56                  & 2.93          & 1.62          & 1.42          & 1.57          \\
                                     & $1 \times 10^{21}$ & 1.55                  & 2.83          & 1.57          & 1.32          & 1.55          \\
                                     & $3 \times 10^{21}$ & 1.48                  & \textbf{2.48} & 1.46          & 1.19          & 1.51          \\ \cmidrule(l){1-2}
\multirow{6}{*}{DoReMi Proportional} & $6 \times 10^{19}$ & 1.72                  & 4.16          & 1.99          & 2.84          & 1.68          \\
                                     & $1 \times 10^{20}$ & 1.68                  & 3.86          & 1.92          & 2.69          & 1.66          \\
                                     & $3 \times 10^{20}$ & 1.61                  & 3.42          & 1.72          & 2.08          & 1.60          \\
                                     & $6 \times 10^{20}$ & 1.57                  & 2.95          & 1.64          & 1.40          & 1.58          \\
                                     & $1 \times 10^{21}$ & 1.56                  & 2.81          & 1.57          & 1.34          & 1.56          \\
                                     & $3 \times 10^{21}$ & 1.48                  & 2.54          & 1.47          & 1.20          & 1.51          \\ \cmidrule(l){1-2}
\multirow{6}{*}{DoReMi Uniform}      & $6 \times 10^{19}$ & 2.38                  & 7.77          & 3.61          & 6.88          & 1.90          \\
                                     & $1 \times 10^{20}$ & 2.32                  & 8.35          & 3.37          & 7.00          & 1.92          \\
                                     & $3 \times 10^{20}$ & 2.18                  & 7.04          & 2.89          & 5.82          & 1.86          \\
                                     & $6 \times 10^{20}$ & 2.07                  & 6.96          & 2.72          & 4.69          & 1.84          \\
                                     & $1 \times 10^{21}$ & 1.99                  & 6.46          & 2.59          & 4.40          & 1.81          \\
                                     & $3 \times 10^{21}$ & 1.96                  & 6.02          & 2.42          & 3.74          & 1.77          \\ \cmidrule(l){1-2}
\multirow{6}{*}{ODM GitHub}          & $6 \times 10^{19}$ & 1.66                  & 4.84          & 1.62          & 2.41          & 1.66          \\
                                     & $1 \times 10^{20}$ & 1.64                  & 4.51          & 1.56          & 2.29          & 1.64          \\
                                     & $3 \times 10^{20}$ & 1.57                  & 3.93          & 1.32          & 1.69          & 1.60          \\
                                     & $6 \times 10^{20}$ & 1.55                  & 3.44          & 1.28          & 1.34          & 1.56          \\
                                     & $1 \times 10^{21}$ & 1.51                  & 3.32          & 1.22          & 1.24          & 1.52          \\
                                     & $3 \times 10^{21}$ & 1.44                  & 2.84          & 1.14          & 1.11          & 1.47          \\ \cmidrule(l){1-2}
\multirow{6}{*}{ODM Paper}           & $6 \times 10^{19}$ & 1.72                  & 5.23          & 1.68          & 2.35          & 1.72          \\
                                     & $1 \times 10^{20}$ & 1.72                  & 4.93          & 1.60          & 2.17          & 1.69          \\
                                     & $3 \times 10^{20}$ & 1.61                  & 4.30          & 1.37          & 1.63          & 1.65          \\
                                     & $6 \times 10^{20}$ & 1.58                  & 3.79          & 1.35          & 1.27          & 1.62          \\
                                     & $1 \times 10^{21}$ & 1.55                  & 3.63          & 1.28          & 1.21          & 1.59          \\
                                     & $3 \times 10^{21}$ & 1.48                  & 3.01          & 1.16          & 1.08          & 1.53          \\ \cmidrule(l){1-2}
\multirow{6}{*}{OLMo Sampling}       & $6 \times 10^{19}$ & 1.65                  & 4.78          & 1.69          & 2.26          & 1.63          \\
                                     & $1 \times 10^{20}$ & 1.60                  & 4.33          & 1.62          & 2.18          & 1.60          \\
                                     & $3 \times 10^{20}$ & 1.55                  & 3.66          & 1.40          & 1.62          & 1.55          \\
                                     & $6 \times 10^{20}$ & 1.51                  & 3.14          & 1.34          & 1.17          & 1.51          \\
                                     & $1 \times 10^{21}$ & 1.46                  & 2.98          & 1.29          & 1.12          & 1.50          \\
                                     & $3 \times 10^{21}$ & \textbf{1.40}         & 2.55          & 1.18          & 1.02          & 1.45          \\ \cmidrule(l){1-2}
\multirow{6}{*}{Greedy Ablation}     & $6 \times 10^{19}$ & 1.65                  & 4.69          & 1.68          & 2.15          & 1.64          \\
                                     & $1 \times 10^{20}$ & 1.64                  & 4.44          & 1.62          & 2.06          & 1.62          \\
                                     & $3 \times 10^{20}$ & 1.58                  & 3.63          & 1.48          & 1.58          & 1.56          \\
                                     & $6 \times 10^{20}$ & 1.53                  & 3.15          & 1.40          & 1.12          & 1.53          \\
                                     & $1 \times 10^{21}$ & 1.49                  & 2.95          & 1.33          & 1.10          & 1.50          \\
                                     & $3 \times 10^{21}$ & 1.43                  & 2.56          & 1.21          & 0.99          & 1.47          \\ \cmidrule(l){1-2}
\multirow{6}{*}{UtiliMax Ablation}   & $6 \times 10^{19}$ & 1.65                  & 4.91          & 1.63          & 2.27          & 1.63          \\
                                     & $1 \times 10^{20}$ & 1.62                  & 4.42          & 1.57          & 2.11          & 1.62          \\
                                     & $3 \times 10^{20}$ & 1.55                  & 3.70          & 1.32          & 1.55          & 1.55          \\
                                     & $6 \times 10^{20}$ & 1.50                  & 3.20          & 1.27          & 1.17          & 1.53          \\
                                     & $1 \times 10^{21}$ & 1.47                  & 3.01          & 1.21          & 1.11          & 1.49          \\
                                     & $3 \times 10^{21}$ & 1.42                  & 2.57          & \textbf{1.13} & 1.01          & 1.45          \\
\end{tabular}
\end{center}
\end{table}

\begin{table}[t]
\caption{(Cont.) NLL Metrics for each individual run with simulated epoching. NLL Per Token is used for MATH, MBPP, Flores. Normalized NLL Per Token is used for ARC Challenge and MMLU, since they are Multiple-Choice Tasks. Best results in \textbf{bold}.}
\label{tab:full-res-2}
\begin{center}
\begin{tabular}{llccccc}
\toprule
\multicolumn{2}{c}{Training Run Setup}                                 & \multicolumn{5}{c}{Downstream Evaluation}   \\
\cmidrule(lr){1-2} \cmidrule(lr){3-7} 
Mixing Method  & FLOPs & ARC & Flores        & MATH          & MBPP    & MMLU         \\ \cmidrule(lr){1-1} \cmidrule(lr){2-2} \cmidrule(lr){3-3} \cmidrule(lr){4-4} \cmidrule(lr){5-5} \cmidrule(lr){6-6} \cmidrule(lr){7-7}
\multirow{6}{*}{Softmax Ablation} & $6 \times 10^{19}$ & 1.90                  & 6.35   & 3.08          & 2.22          & 1.87          \\
                                  & $1 \times 10^{20}$ & 1.87                  & 6.10   & 4.13          & 2.15          & 1.81          \\
                                  & $3 \times 10^{20}$ & 1.81                  & 5.48   & 1.47          & 1.60          & 1.89          \\
                                  & $6 \times 10^{20}$ & 1.76                  & 4.72   & 1.91          & 1.20          & 1.76          \\
                                  & $1 \times 10^{21}$ & 1.72                  & 4.59   & 1.55          & 1.15          & 1.73          \\
                                  & $3 \times 10^{21}$ & 1.67                  & 3.82   & 1.54          & 1.01          & 1.73          \\ \cmidrule(l){1-2}
\multirow{6}{*}{Proportional}     & $6 \times 10^{19}$ & 1.63                  & 4.70   & 1.73          & 2.42          & 1.63          \\
                                  & $1 \times 10^{20}$ & 1.59                  & 4.22   & 1.66          & 2.28          & 1.60          \\
                                  & $3 \times 10^{20}$ & 1.55                  & 3.61   & 1.44          & 1.67          & 1.56          \\
                                  & $6 \times 10^{20}$ & 1.51                  & 3.19   & 1.38          & 1.20          & 1.53          \\
                                  & $1 \times 10^{21}$ & 1.48                  & 2.92   & 1.31          & 1.17          & 1.50          \\
                                  & $3 \times 10^{21}$ & 1.41                  & 2.52   & 1.22          & 1.07          & 1.44          \\ \cmidrule(l){1-2}
\multirow{6}{*}{Greedy MEDU}      & $6 \times 10^{19}$ & 1.71                  & 5.29   & 1.63          & 2.16          & 1.66          \\
                                  & $1 \times 10^{20}$ & 1.63                  & 4.95   & 1.57          & 2.05          & 1.61          \\
                                  & $3 \times 10^{20}$ & 1.57                  & 4.24   & 1.32          & 1.53          & 1.57          \\
                                  & $6 \times 10^{20}$ & 1.52                  & 3.40   & 1.28          & 1.09          & 1.52          \\
                                  & $1 \times 10^{21}$ & 1.48                  & 3.28   & 1.21          & 1.04          & 1.49          \\
                                  & $3 \times 10^{21}$ & 1.42                  & 2.71   & \textbf{1.13} & \textbf{0.94} & 1.45          \\ \cmidrule(l){1-2}
\multirow{6}{*}{UtiliMax MEDU}    & $6 \times 10^{19}$ & 1.65                  & 4.77   & 1.63          & 2.14          & 1.64          \\
                                  & $1 \times 10^{20}$ & 1.62                  & 4.49   & 1.57          & 2.08          & 1.61          \\
                                  & $3 \times 10^{20}$ & 1.56                  & 3.72   & 1.33          & 1.51          & 1.55          \\
                                  & $6 \times 10^{20}$ & 1.51                  & 3.18   & 1.27          & 1.14          & 1.52          \\
                                  & $1 \times 10^{21}$ & 1.49                  & 3.00   & 1.22          & 1.09          & 1.50          \\
                                  & $3 \times 10^{21}$ & 1.41         & 2.50   & 1.13          & 1.00          & 1.45 \\ \cmidrule(l){1-2}
\multirow{6}{*}{Softmax MEDU}     & $6 \times 10^{19}$ & 1.73                  & 5.43   & 2.84          & 2.34          & 1.94          \\
                                  & $1 \times 10^{20}$ & 1.71                  & 5.04   & 2.94          & 2.17          & 1.81          \\
                                  & $3 \times 10^{20}$ & 1.65                  & 4.46   & 2.89          & 1.64          & 1.84          \\
                                  & $6 \times 10^{20}$ & 1.58                  & 3.95   & 1.72          & 1.26          & 1.75          \\
                                  & $1 \times 10^{21}$ & 1.60                  & 3.70   & 1.38          & 1.19          & 1.60          \\
                                  & $3 \times 10^{21}$ & 1.51                  & 3.13   & 1.16          & 1.06          & 1.54          \\ \cmidrule(l){1-2}
\multirow{6}{*}{UniMax}           & $6 \times 10^{19}$ & 1.63                  & 4.87   & 1.63          & 2.22          & 1.63          \\
                                  & $1 \times 10^{20}$ & 1.63                  & 4.36   & 1.57          & 2.11          & 1.62          \\
                                  & $3 \times 10^{20}$ & 1.57                  & 3.62   & 1.32          & 1.52          & 1.56          \\
                                  & $6 \times 10^{20}$ & 1.51                  & 3.19   & 1.28          & 1.18          & 1.52          \\
                                  & $1 \times 10^{21}$ & 1.48                  & 2.98   & 1.22          & 1.12          & 1.50          \\
                                  & $3 \times 10^{21}$ & 1.42                  & 2.54   & 1.13          & 1.01          & \textbf{1.44} \\ \cmidrule(l){1-2}
\multirow{6}{*}{Uniform}          & $6 \times 10^{19}$ & 1.73                  & 5.32   & 1.69          & 2.31          & 1.71          \\
                                  & $1 \times 10^{20}$ & 1.69                  & 4.85   & 1.61          & 2.18          & 1.70          \\
                                  & $3 \times 10^{20}$ & 1.63                  & 4.28   & 1.40          & 1.63          & 1.65          \\
                                  & $6 \times 10^{20}$ & 1.59                  & 3.77   & 1.36          & 1.27          & 1.61          \\
                                  & $1 \times 10^{21}$ & 1.55                  & 3.62   & 1.25          & 1.17          & 1.57          \\
                                  & $3 \times 10^{21}$ & 1.49                  & 3.02   & 1.18          & 1.06          & 1.53          \\ \bottomrule
\end{tabular}
\end{center}
\end{table}

\end{document}